\documentclass[acmsmall]{acmart}
\AtBeginDocument{%
  }

\setcopyright{acmlicensed}
\copyrightyear{2025}
\acmYear{2025}
\acmDOI{XXXXXXX.XXXXXXX}

\usepackage{multirow}%
\definecolor{gold}{rgb}{1.0, 0.84, 0.0}       
\definecolor{lightskyblue}{rgb}{0.53, 0.81, 0.98} 
\definecolor{lightsalmon}{rgb}{1.0, 0.63, 0.48}  
\begin{document}

\title{Leveraging Auxiliary Information in Text-to-Video Retrieval: A Review}

\author{Adriano Fragomeni}
\email{adriano.fragomeni@bristol.ac.uk}
\affiliation{%
  \institution{University of Bristol}
  \city{Bristol}
  \country{UK}
}

\author{Dima Damen}
\email{dima.damen@bristol.ac.uk}
\affiliation{%
  \institution{University of Bristol}
  \city{Bristol}
  \country{UK}
}

\author{Michael Wray}
\email{michael.wray@bristol.ac.uk}
\affiliation{%
  \institution{University of Bristol}
  \city{Bristol}
  \country{UK}
}

\renewcommand{\shortauthors}{Fragomeni et al.}

\begin{abstract}
Text-to-Video (T2V) retrieval aims to identify the most relevant item from a gallery of videos based on a user’s text query. 
Traditional methods rely solely on aligning video and text modalities to compute the similarity and retrieve relevant items. However, recent advancements emphasise incorporating auxiliary information extracted from video and text modalities to improve retrieval performance and bridge the semantic gap between these modalities. Auxiliary information can include visual attributes, such as objects; temporal and spatial context; and textual descriptions, such as speech and rephrased captions. 

This survey comprehensively reviews 81 research papers on Text-to-Video retrieval that utilise such auxiliary information. It provides a detailed analysis of their methodologies; highlights state-of-the-art results on benchmark datasets; and discusses available datasets and their auxiliary information. Additionally, it proposes promising directions for future research, focusing on different ways to further enhance retrieval performance using this information.
\end{abstract}


\received{30 May 2025}

\maketitle

\section{Introduction}\label{sec1}
The increasing number of videos uploaded on platforms such as YouTube, TikTok, and Instagram has established video as a dominant medium for communication and entertainment. However, this rapid growth presents significant challenges for efficiently retrieving relevant content given a user’s text query. Traditional video retrieval methods struggle to align video and text modalities and address the complexity of videos because these methods typically rely on limited textual queries that are not able to capture the rich visual content present in a video.

Video retrieval involves aligning video and text data by processing video-text pairs to extract relevant representations. Videos, with their dynamic visuals, temporal structures, and audio signals, are completely different from the structured grammar and semantics of text, making this a challenging multimodal problem. Bridging this gap requires algorithms that capture these relationships, surpassing the limitations of single-modal methods like image classification or text search.

The Text-to-Video retrieval task can be divided into two main categories: Sentence-to-Clip (S2C) and Paragraph-to-Video (P2V) retrieval. Sentence-to-Clip focuses on retrieving short clips matching specific text queries and requires fine-grained alignment of actions or objects. For example, a user searching for \emph{``a cat jumping on a table''} would expect the model to be able to distinguish the jumping action from other actions and the cat from other animals. In contrast, Paragraph-to-Video involves retrieving entire videos based on longer, multi-sentence text queries, reflecting multiple actions in the video content. For instance, a user might search for \emph{``tutorial of making pasta. boiling water. Put pasta in the pot. Mix pasta with sauce. Serving pasta.''}, requiring the model to retrieve a video that aligns with the sequence of actions described in the query.

Despite progress in Text-to-Video retrieval, many challenges remain due to the multimodal complexity and the semantic gap between video and text data. Recent advancements explore integrating auxiliary information—such as object annotations, audio, temporal labels, and additional captions—to enhance retrieval architectures. This enrichment aims to improve alignment between modalities and relevance in retrieval output.

In the following sections, we will analyse the evolution of video retrieval methods that integrate auxiliary information to improve the retrieval performance. By examining these strategies, we aim to show the importance of this extra information in Text-to-Video retrieval by providing a comprehensive view of methods and benchmark datasets that incorporate this information and potential future directions on how to use such information in this task.

\begin{figure}[ht!]
    \centering
    \includegraphics[width=\textwidth]{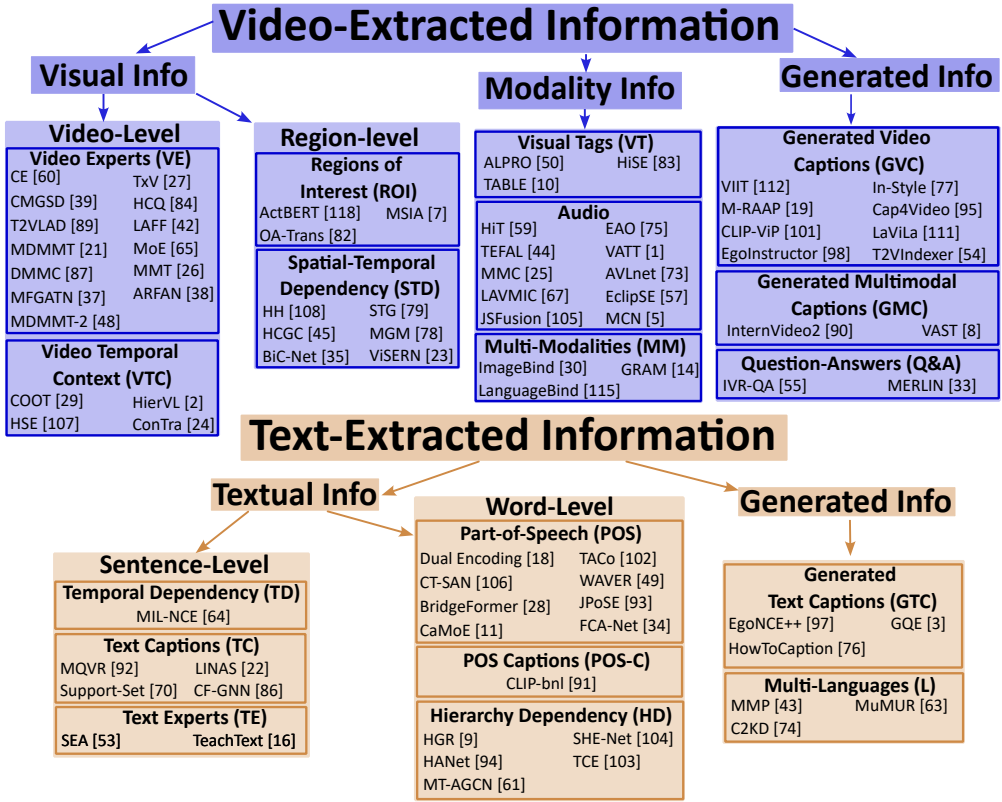}
    \caption{Overview of different auxiliary information extracted from video (\textcolor{blue}{top}) and text (\textcolor{orange}{bottom}) used in Text-to-Video retrieval. Auxiliary information from video can be divided into Visual Info (Video-Level and Region-Level), Modality Info and Generated Info. On the contrary, auxiliary information from text can be Textual Info (Sentence-Level and Word-Tagging) or Generated Info.}
    \label{fig:additional_information}
\end{figure}

\subsection{Motivation}
Text-to-Video retrieval faces a fundamental challenge: bridging the gap between two distinct modalities—text (symbolic and structured) and video (dynamic and diverse). These modalities differ significantly in how they represent information. Text relies on grammar and semantics to communicate ideas, while video captures dynamic visual and temporal content that is naturally complex and diverse. The task of aligning these two modalities has driven the development of numerous methods in Text-to-Video retrieval.

A growing number of approaches aim to overcome this problem by incorporating auxiliary information extracted from videos and texts, as shown in Fig.~\ref{fig:additional_information}. Auxiliary information from video is divided into visual, modality and generated info, whereas auxiliary information from text is divided into textual and generated. Visual info is that type of information that belongs to the video modality, such as Region of Interests (ROIs); modality info consists of information that belongs to different modalities from the video, such as text, audio, infrared; textual info belongs to the text modality, such as Part-Of-Speech (POS) tags; and generated info is information that is generated by Visual-Language Models (VLMs) and Large Language Models (LLMs). Overall, this auxiliary information has proven to be an important tool for bridging the semantic gap between text and video. For instance, audio can capture spoken phrases or background sounds, such as stadium cheers, which enrich the contextual understanding of the video.

Despite this, there is a significant gap in the literature regarding reviewing these methods.
While a few surveys have summarised general developments in Text-to-Video retrieval, none have focused on a detailed analysis of works that leverage this auxiliary information. Given that auxiliary information represents one of the potential futures of this field, with new methods increasingly utilising it to boost retrieval performances, we believe providing a detailed review is essential. Understanding how auxiliary information can be integrated and how these approaches compare across benchmark datasets is essential to guide future research and provide better clarity in this rapidly evolving research domain. This survey aims to fill this void by reviewing these works, offering a deep analysis and classification of them.
The key motivations are:
    (1) \textbf{Addressing Gaps in Literature}: Existing surveys overlook methods using auxiliary information. This survey focuses exclusively on these approaches.
    (2) \textbf{Structured Analysis}: We categorise auxiliary information into information extracted from video (video-extracted) and text (text-extracted).
    (3) \textbf{Dataset Review}: A detailed analysis of benchmark datasets offering auxiliary information for retrieval.
    (4)  \textbf{Comparative Evaluation}: We compare methods leveraging auxiliary information, ranking the various approaches.
    (5) \textbf{Future Directions}: The survey identifies possible directions for leveraging auxiliary data in Text-to-Video retrieval.
By achieving these objectives, this survey serves as a comprehensive resource for advancing research in Text-to-Video retrieval.

\subsection{Related Surveys}

Cross-modal retrieval has been widely studied, with surveys covering the tasks of text-image, text-video (i.e. Sentence-to-Clip and Paragraph-to-Video approaches), and general retrieval (i.e. image and video approaches). These works offer insights into methodologies, challenges, datasets, and evaluation metrics, and can be divided into three groups:

\begin{enumerate}
\item \textbf{Text-image retrieval} explores the alignment of text and image data. Surveys such as~\cite{DBLP:conf/ijcai/CaoLLNZ22} review traditional and deep learning approaches, addressing challenges like dataset limitations and the influence of large-scale pre-trained models. Similarly,~\cite{li2024bridging} highlights advancements in Multimodal Large Language Models (MLLMs), categorising retrieval methods into single-stream, dual-stream, and hashing methods, emphasising generalisation challenges.

\item \textbf{Text-video retrieval} introduces complexity in the retrieval task due to the temporal and multimodal nature of videos. Early surveys like~\cite{DBLP:journals/corr/abs-1205-1641} discuss feature extraction and dimensionality reduction, while recent surveys such as~\cite{DBLP:journals/corr/abs-2405-03770} categorise models into image-based, video-based, and multimodal types. Advances in deep learning, such as transformers and contrastive losses, are reviewed in~\cite{DBLP:journals/ijmir/ZhuJCGL23}, which also addresses computational efficiency. Further innovations, including self-attention mechanisms and multi-task learning, are analysed in~\cite{DBLP:journals/mta/WuQW23}. Expanding this,~\cite{DBLP:conf/acl/NguyenBXQ0WNNL24} explores video-language tasks, identifying challenges in retrieval, captioning, and question answering.

\item \textbf{General retrieval} provides an overview of the retrieval task across different modalities. Foundational techniques such as statistical and deep learning based approaches are reviewed in~\cite{DBLP:journals/corr/WangY0W016,DBLP:journals/access/HanAMK24}, focusing on semantic gap reduction and scalability. Unified frameworks for integrating multiple modalities are analysed in~\cite{DBLP:journals/tcsv/PengHZ18}, while~\cite{DBLP:journals/csr/KaurPM21} evaluates cross-modal methods on benchmarks, focusing on how to bridge the gap between different modalities. In contrast, our review diverges from recent works that primarily focus on efficient multimodal hashing~\cite{DBLP:journals/tkde/ZhuZGLYS24} or on deep learning advances in video retrieval that consider only object detectors as auxiliary information~\cite{DBLP:journals/access/ZhouHH23}, omitting all the other other auxiliary information available from videos and texts. Additionally, unlike~\cite{DBLP:journals/corr/abs-2308-14263}, which solely focuses on retrieval tasks involving a wide range of data modalities such as video-audio and text-3D, our work specifically focuses on Text-to-Video retrieval involving only text-video pairs as primary inputs.
\end{enumerate}

While the reviewed surveys collectively highlight diverse cross-modal approaches, from text-image to text-video retrieval, they often ignore the added value of auxiliary information, such as external metadata, temporal context, audio modality and additional captions. By focusing specifically on how this auxiliary information extracted or generated from video and text data can be used in Text-to-Video retrieval, our review provides a new, distinct perspective on this task that has never been explored. We analyse and classify 81 approaches based on the type of auxiliary information integrated. In doing so, we suggest possible future directions in Text-to-Video retrieval by highlighting how the integration of auxiliary information can evolve this task further.

\subsection{Article Organization}
The review is structured as follows: Section~\ref{sec3} introduces an overview of all methods utilising auxiliary information extracted from videos; Section~\ref{sec4} explores approaches leveraging auxiliary information extracted from texts; Section~\ref{sec5} reviews benchmark datasets and their available auxiliary information, comparing results; Section~\ref{sec6} discusses research directions for leveraging auxiliary information in retrieval tasks; and Section~\ref{sec7} concludes this review paper.

\section{Auxiliary Information from Video}\label{sec3}
This section explores methods leveraging additional video-extracted information, focusing on how it is incorporated into the proposed architectures. Techniques using visual info are discussed in Sec.~\ref{subsec3:visual}, methods employing modality info, e.g. text and audio, in Sec.~\ref{subsec3:textual_audio}, and approaches utilising generative models to produce auxiliary information, i.e. generated info, in Sec.~\ref{subsec3:visual_generated}. Tab.~\ref{tab:explanation_auxiliary_video} shows the categorisation of this auxiliary information by also giving a general definition for each one.

\begin{table*}[t!]
\centering
\resizebox{\linewidth}{!}{%
\begin{tabular}{llll}

& &\textbf{Class}
& \textbf{Auxiliary Information} \\
\hline

\multirow{4}{*}{\centering Visual Info}&\multirow{2}{*}{\centering Video-Level} & Video Experts (VE)
 & Information extracted by specialised pre-trained models.\\
 
 && Video Temporal Context (VTC)
 & Temporal information derived from neighbouring clips.\\\cline{2-4}

&\multirow{2}{*}{\centering Region-Level}
 & Regions of Interest (ROI)
 & Bounding boxes of important regions in anchor frames.\\

 && Spatial-Temporal Dependency (STD)
 & Spatial and temporal relationships among ROIs. \\\hline

\multirow{3}{*}{\centering Modality Info}&&Video Tags (VT)& Keywords summarising or categorising video content.\\

&&Audio&  Speech and non-speech sounds, e.g. ambient noise.\\

&&Multi-Modalities (MM)& Additional video modalities, e.g. infrared.\\\hline

\multirow{3}{*}{\centering Generated Info}&&Generated Video Captions (GVC)& Additional captions generated from the video.\\

&&Generated Multimodal Captions (GMC)& Additional captions generated from additional modalities.\\

&&Questions $\&$ Answers (Q$\&$A)& Interactive question-answer system.\\\hline\hline
\end{tabular}}
\caption{Overview of video-extracted auxiliary information, divided into Visual, Modality and Generated Info. Visual Info is further divided into Video-Level/Region-Level. Each
category includes specific auxiliary classes.}
\label{tab:explanation_auxiliary_video}
\end{table*}

\subsection{Visual Info}\label{subsec3:visual}
Videos are multimodal data, containing not only RGB information but also additional visual information that can enhance video retrieval performance.
Different approaches have utilised this video-extracted information in video retrieval. These works can be categorised into Video-Level and Region-Level. Video-Level methods use auxiliary information extracted globally from videos, whereas Region-Level methods use local information extracted from individual frames.

\subsubsection{Video-Level Methods} 
Video-Level methods leverage diverse visual information such as Video Experts (VE), which include different video domain-specific embeddings~\cite{DBLP:journals/corr/abs-1804-02516,DBLP:conf/bmvc/LiuANZ19,DBLP:conf/eccv/Gabeur0AS20,DBLP:conf/cvpr/WangZ021,DBLP:conf/cvpr/DzabraevKKP21,DBLP:journals/corr/abs-2203-07086,DBLP:conf/sigir/HeWFJLZT21,DBLP:conf/ijcai/WangZCCZPGWS21, DBLP:conf/icmcs/HaoZWZLWM21,DBLP:conf/mir/HaoZWZLW21,DBLP:conf/www/WangCLZLXX22,DBLP:conf/eccv/GalanopoulosM22, DBLP:conf/eccv/HuCWZDL22} and Video Temporal Context (VTC), which involves neighbouring clips~\cite{DBLP:conf/eccv/ZhangHS18,DBLP:conf/nips/GingZPB20,DBLP:conf/accv/FragomeniWD22,DBLP:conf/cvpr/AshutoshGTG23}.

\noindent\textbf{Video Experts (VE)} consists of different video embeddings that can be combined to enrich the final video representation. These embeddings are domain-specific modalities such as Appearance, Face, Hands, Motion, Optical Character Recognition (OCR), and Scene, which represent different aspects of the same video, or coarse video features extracted from pre-trained architectures. Each expert is always treated as a separate input to the model.

Many works propose to use different domain-specific embeddings to improve video retrieval~\cite{DBLP:journals/corr/abs-1804-02516,DBLP:conf/bmvc/LiuANZ19,DBLP:conf/eccv/Gabeur0AS20,DBLP:conf/cvpr/DzabraevKKP21,DBLP:journals/corr/abs-2203-07086,DBLP:conf/ijcai/WangZCCZPGWS21,DBLP:conf/sigir/HeWFJLZT21,DBLP:conf/www/WangCLZLXX22,DBLP:conf/mir/HaoZWZLW21,DBLP:conf/icmcs/HaoZWZLWM21}. The Mixture of Experts model (MoE)~\cite{DBLP:journals/corr/abs-1804-02516} proposes to learn joint text-video embeddings by combining multiple modality-specific video experts with a separate text embedding learnt for each of them. The final similarity score between video and text is computed as a weighted sum of expert similarity scores, where the weights are estimated based on the input text. Similarly, Collaborative Experts (CE)~\cite{DBLP:conf/bmvc/LiuANZ19} uses various pre-trained modality-specific experts (e.g. motion, appearance, scene, OCR, faces, audio, and ASR) to construct enriched video representations. A dynamic attention mechanism prioritises relevant embeddings and enables expert-level interactions. These embeddings are fused into a fixed-length representation optimised for video-text retrieval. 

Building on these two works, MMT~\cite{DBLP:conf/eccv/Gabeur0AS20} introduces a transformer-based architecture that encodes multiple video experts and captures both cross-modal relationships and long-term temporal patterns. Many of the following papers~\cite{DBLP:journals/corr/abs-2203-07086,DBLP:conf/cvpr/DzabraevKKP21,DBLP:conf/ijcai/WangZCCZPGWS21,DBLP:conf/sigir/HeWFJLZT21,DBLP:conf/www/WangCLZLXX22} build upon its architecture. MDMMT~\cite{DBLP:conf/cvpr/DzabraevKKP21} and MDMMT-2~\cite{DBLP:journals/corr/abs-2203-07086} integrate multimodal transformers with extensive multi-dataset training. DMMC~\cite{DBLP:conf/ijcai/WangZCCZPGWS21} employs a hierarchical attention mechanism that aligns local (word-video) and global (sentence-video) representations using modality-specific transformers followed by a holistic transformer for high-level semantics. CMGSD~\cite{DBLP:conf/sigir/HeWFJLZT21} incorporates an adaptive triplet loss margin based on negative pair similarity and applies a Cross-Modal Generalised Self-Distillation mechanism to refine alignment. HCQ~\cite{DBLP:conf/www/WangCLZLXX22} introduces a Hybrid Contrastive Quantisation technique to perform multi-level quantised embeddings for each modality to balance storage efficiency and retrieval performance. Inspired by DMMC, T2VLAD~\cite{DBLP:conf/cvpr/WangZ021} combines global and local alignment to capture details at different levels of granularity.

Alternatively, graph-based aggregation strategies have also been explored~\cite{DBLP:conf/mir/HaoZWZLW21,DBLP:conf/icmcs/HaoZWZLWM21} to process different video experts. MFGATN~\cite{DBLP:conf/mir/HaoZWZLW21} uses a Multi-Feature Graph Attention Network to model relationships among video experts as graph nodes. It applies graph attention layers to capture high-level semantic relations between modalities. Similarly, ARFAN~\cite{DBLP:conf/icmcs/HaoZWZLWM21} employs a self-attention mechanism to assign higher weights to the most representative experts based on query relevance.

In contrast to these approaches, a few methods focus on aggregating coarse-grained visual features extracted from different pre-trained video models~\cite{DBLP:conf/eccv/GalanopoulosM22,DBLP:conf/eccv/HuCWZDL22}. TxV~\cite{DBLP:conf/eccv/GalanopoulosM22} employs multiple pre-trained networks to extract diverse representations of the same video, mapping each into a separate latent space.
Similarly, LAFF~\cite{DBLP:conf/eccv/HuCWZDL22} applies an attention mechanism to multiple video features, allowing the model to learn which of them is more important for matching with a text query. Multiple parallel LAFF modules capture different aspects or relationships among the video features and the text.

\textbf{Video Temporal Context (VTC)} enriches video features by incorporating temporal information from neighbouring clips, producing temporally enriched video embeddings.

HSE~\cite{DBLP:conf/eccv/ZhangHS18} introduces a Hierarchical Sequence Embeddings to address Sequence-to-Clip (S2C) and Paragraph-to-Video (P2V) relations. The model aligns clip-sentence pairs locally and video-paragraph pairs globally. Building on this concept, COOT~\cite{DBLP:conf/nips/GingZPB20} advances hierarchical video-text retrieval with a multi-level transformer architecture. Frame- and word-level features of untrimmed videos are encoded with a temporal transformer, while an attention-aware feature aggregation module refines segment-level (clip or sentence) features. Finally, a contextual transformer integrates local and global features into a unified video or paragraph embedding. Following~\cite{DBLP:conf/eccv/ZhangHS18,DBLP:conf/nips/GingZPB20}, which focus on P2V scenarios, ConTra~\cite{DBLP:conf/accv/FragomeniWD22} introduces a transformer-based model for S2C retrieval in untrimmed videos. By leveraging neighbouring clips within a temporal window, ConTra creates contextually enriched video embeddings and employs a neighbouring contrastive loss to ensure distinct representations for overlapping temporal contexts, helping the model to distinguish clips in consecutive sequences. 

More recently, HierVL~\cite{DBLP:conf/cvpr/AshutoshGTG23} proposes a dual-layered hierarchical architecture for egocentric video representations. The first layer aligns action-specific temporal context by aligning each clip with its corresponding sentence. The second layer aggregates short-term features into long-term representations aligned with the overall narrative.

\subsubsection{Region-Level} 
This group of works focuses on using Regions of Interest (ROI) as auxiliary information. Specifically, these approaches use object detectors to extract key objects or regions within frames. Some methods rely exclusively on ROIs extracted from an anchor frame, which is the most representative frame in a video~\cite{DBLP:conf/cvpr/ZhuY20a,DBLP:conf/cvpr/WangGCY0SQS22,chen2024multilevel}, others use more frames of a video to consider also their Spatial-Temporal Dependency (STD)~\cite{DBLP:conf/ijcai/FengZG020,DBLP:conf/iccv/Zhang0Z23a,DBLP:conf/mm/SongCJ23,DBLP:journals/tomccap/HanZSXCC24,DBLP:conf/sigir/JinZZZHZ21,DBLP:journals/tmm/SongCWJ22}.

\textbf{Regions of Interest (ROI)} enhance video-text alignment by focusing on key objects or regions extracted from an anchor frame, prioritising static visual features over temporal dynamics to simplify Text-to-Video retrieval.

One example is ActBERT~\cite{DBLP:conf/cvpr/ZhuY20a}, which employs a Transformer to align global action representations with localised ROIs, whose features are extracted from bounding boxes, and associated textual descriptions. In this approach, ROIs are extracted from an anchor frame, while temporal information is also modelled using multiple clips for long-term video context understanding. OA-Trans~\cite{DBLP:conf/cvpr/WangGCY0SQS22} proposes an object-aware transformer to incorporate object-level features from anchor frames into a dual-encoder architecture. The model employs an object-guided masking technique to exclude non-object regions and preserves only the most salient objects for alignment with the text input. Similarly, MSIA~\cite{chen2024multilevel} identifies and aligns key regions, extracted from an anchor frame, with nouns and phrases in the corresponding textual descriptions. 

\textbf{Spatial-Temporal Dependency (STD)} enhances video retrieval tasks by integrating ROIs extracted from multiple frames and leveraging their spatial-temporal dynamics.

Several studies employ graph-based methods~\cite{DBLP:conf/ijcai/FengZG020,DBLP:conf/sigir/JinZZZHZ21,DBLP:journals/tmm/SongCWJ22} to model spatial-temporal relationships among ROIs within and across frames by using graph structures.
For example, a graph-based framework, where ROIs serve as vertices and semantic relations as edges, is introduced in~\cite{DBLP:conf/ijcai/FengZG020}, where a Graph Convolutional Network (GCN), combined with a random walk-based reasoning, enhances region-level features, reduces redundancy, and enriches spatial reasoning at the frame level. 
HCGC~\cite{DBLP:conf/sigir/JinZZZHZ21} extends this idea by constructing hierarchical graphs for both visual and textual modalities, where objects and actions are represented as graph nodes to model their spatial-temporal dependencies. Similarly, STG~\cite{DBLP:journals/tmm/SongCWJ22} represents videos as spatial-temporal graphs, with objects as nodes and their interactions as edges. 

Other works leverage this spatial-temporal dependency using transformer architectures~\cite{DBLP:conf/mm/SongCJ23,DBLP:conf/iccv/Zhang0Z23a,DBLP:journals/tomccap/HanZSXCC24}.
MGM~\cite{DBLP:conf/mm/SongCJ23} represents video content through subject-verb-object (SVO) triplets, where ROIs represent objects in frames, and their relationships form triplets that capture their spatial-temporal evolution. These triplets are aligned with textual descriptions to model both static interactions and their temporal dynamics in a fine-grained manner. In parallel, the model computes a coarse-grained alignment between the overall video and text embedding. The final retrieval decision combines both local (triplet-level) and global (video-level) matching information.
BiC-Net~\cite{DBLP:journals/tomccap/HanZSXCC24} uses a dual-branch approach to model local and global temporal patterns. A Spatio-Temporal Residual Transformer captures detailed object relations, while a global branch extracts global temporal information.

Different from previous works, Helping Hands (HH)~\cite{DBLP:conf/iccv/Zhang0Z23a} focuses on egocentric videos and proposes to predict hand and object positions with their semantic categories during training. ROIs are detected by an off-the-shelf hand object detector and integrated into a transformer architecture to model their spatial-temporal dependencies.

\subsection{Modality Info}\label{subsec3:textual_audio}
These methods can be grouped based on the extra modalities used in addition to the visual information, such as Video Tags (VT)—i.e. textual labels or keywords describing a video’s topic or visual key elements---~\cite{DBLP:conf/cvpr/Li0LNH22,DBLP:conf/mm/WangXHLJHD22,DBLP:conf/aaai/ChenWLQMS23}, Audio modality---i.e. speech and sounds~\cite{DBLP:conf/eccv/YuKK18,DBLP:conf/interspeech/RouditchenkoBHC21,DBLP:conf/wacv/GabeurN0AS22,DBLP:conf/iccv/ChenRDK0BPKFHGP21,DBLP:conf/iccv/Liu0QCDW21,DBLP:conf/nips/AkbariYQCCCG21,DBLP:conf/cvpr/ShvetsovaCR0KFH22,DBLP:conf/eccv/LinLBB22,DBLP:conf/eccv/NagraniSSHMSS22,DBLP:conf/iccv/IbrahimiSWGSO23,DBLP:conf/cvpr/GirdharELSAJM23} or Multi-Modalities (MM)~\cite{DBLP:conf/iclr/ZhuLNYCWPJZLZ0024,DBLP:conf/cvpr/GirdharELSAJM23,DBLP:conf/iclr/ZhuLNYCWPJZLZ0024}---e.g. infrared, depth, and IMU.

\textbf{Video Tags (VT)} are leveraged to improve the alignment between video and text modalities by integrating semantic textual tags extracted from videos using multiple pre-trained experts.
In~\cite{DBLP:conf/cvpr/Li0LNH22}, ALPRO~\cite{DBLP:conf/cvpr/Li0LNH22}, instead of relying on predefined object labels, generates entity prompts based on frequent nouns in the training data. Its main contribution is the Prompting Entity Modelling objective, which predicts entities in randomly sampled video segments to improve region-entity alignment. This objective is trained jointly with a standard video-text contrastive loss in a multimodal transformer architecture.
HiSE~\cite{DBLP:conf/mm/WangXHLJHD22} incorporates Visual Tags by decomposing them into discrete tags, which capture localised objects and actions, and holistic tags, which reflect the overall scene or activity semantics. A graph reasoning module models tag relationships, and the aggregated tag information is integrated into the final video representation. A similar tag-enhanced pipeline is applied to the textual modality. A similar pipeline is proposed for the text component. Finally, TABLE~\cite{DBLP:conf/aaai/ChenWLQMS23} uses Visual Tags as semantic anchors for cross-modal alignment. These tags are extracted from various pre-trained experts, such as objects, scenes, motions, and audio. A cross-modal transformer encoder integrates these tags into both video and text representations. In addition to contrastive learning, Video-Text Matching (VTM) and Masked Language Modelling (MLM) objectives are employed to refine alignment and enhance cross-modal interaction.

\textbf{Audio} captures additional details that video or text alone might miss or cannot express. For example, in a cooking video, the sound of frying onions offers important insights that cannot be inferred just by looking at the video or reading the corresponding text description. Consequently, retrieval methods can better match the text query to the video content, effectively bridging the gap between these two modalities.

Early approaches like JSFusion~\cite{DBLP:conf/eccv/YuKK18} introduce a joint semantic tensor to align video and text; audio contributes indirectly via hierarchical attention mechanisms. However, due to the lack of dedicated audio processing, the impact of audio remains limited.
The following works leverage self-supervised and contrastive learning approaches to integrate audio more effectively~\cite{DBLP:conf/interspeech/RouditchenkoBHC21,DBLP:conf/wacv/GabeurN0AS22,DBLP:conf/iccv/ChenRDK0BPKFHGP21}. 
AVLnet~\cite{DBLP:conf/interspeech/RouditchenkoBHC21} introduces a three-branch model that processes audio, text and video separately. Pre-trained encoders extract features from all the modalities, which are then aligned in a shared embedding space using contrastive objectives. Complementing this idea, MMC~\cite{DBLP:conf/wacv/GabeurN0AS22} presents a masking-reconstruction strategy: one modality (e.g. audio) is masked and reconstructed from the other ones (e.g. video and text).
MCN~\cite{DBLP:conf/iccv/ChenRDK0BPKFHGP21} introduces a clustering-based approach that applies K-means to mean-pooled features from the video, text, and audio modalities. In the shared embedding space, each centroid is an anchor to pull the tri-modal features closer together based on their proximity to the corresponding centroid, which represents the fused video-text-audio features.

Other methods explore Transformers~\cite{DBLP:conf/iccv/Liu0QCDW21,DBLP:conf/nips/AkbariYQCCCG21,DBLP:conf/cvpr/ShvetsovaCR0KFH22,DBLP:conf/iccv/IbrahimiSWGSO23,DBLP:conf/eccv/LinLBB22}, for example, HiT~\cite{DBLP:conf/iccv/Liu0QCDW21} proposes a hierarchical transformer, applying cross-modal contrastive matching at both low-level features and high-level semantics. Audio is fused with appearance and motion to create the final video representation. Similarly, VATT~\cite{DBLP:conf/nips/AkbariYQCCCG21} encodes audio, video, and text inputs via modality-specific projections before processing them jointly using transformers. Additionally,
EAO~\cite{DBLP:conf/cvpr/ShvetsovaCR0KFH22} introduces a fusion transformer to model any combination of video, text and audio modalities, and a combinatorial contrastive loss is applied to learn robust representations across single and paired modalities. Similarly, TEFAL~\cite{DBLP:conf/iccv/IbrahimiSWGSO23} proposes a text-conditioned attention mechanism where audio and video representations are conditioned on the text and then fused via summation to generate the final video representation for both training and evaluation. EclipSE~\cite{DBLP:conf/eccv/LinLBB22} extends the CLIP framework by fusing sparse frames with dense audio information, extracted via convolutional encoders, through cross-modal transformers.

Different from these works, LAVMIC~\cite{DBLP:conf/eccv/NagraniSSHMSS22} proposes to use images in image captioning datasets to mine related audio-visual clips. For each image-caption pair in the dataset, frames in videos with high visual similarity to the image are identified and used to extract short video segments around them. Then, a multimodal transformer-based model is trained on these segments associated with the corresponding captions.

\textbf{Multi-Modalities (MM)} are introduced to learn a unified embedding space where diverse data modalities, mainly extracted from video data, are aligned, showing great performance in many different tasks, including Text-to-Video retrieval. 

ImageBind~\cite{DBLP:conf/cvpr/GirdharELSAJM23} proposes a framework for aligning six different modalities, including visual, audio, text, depth, thermal, and IMU signals, into a shared embedding space using images as the intermediary to align these modalities with language. Audio is embedded through large-scale image-audio web data, and IMU via egocentric video datasets. The architecture consists of modality-specific transformer-based encoders, trained in a contrastive learning setup. Following this work, LanguageBind~\cite{DBLP:conf/iclr/ZhuLNYCWPJZLZ0024} redefines the alignment by adopting language directly as the anchor modality. It unifies multiple modalities, such as audio, image, video, depth, and thermal, in a shared embedding space, with video playing a central role in generating  modalities such as depth and infrared. 

In contrast to these anchor-based approaches, GRAM~\cite{DBLP:conf/iclr/CicchettiGSC25}, instead of aligning each modality to a designated anchor, aligns all modality embeddings jointly by minimising the volume of the parallelotope defined by the modality vectors in a high-dimensional space. This approach enables joint alignment across all modalities, capturing both anchor and non-anchor relationships.

\subsection{Generated Info}\label{subsec3:visual_generated}
Text-to-Video retrieval has shown a large advancement with the introduction of generative models in the retrieval pipeline. These models have been proposed to create new auxiliary data from the video modalities. Common outputs include Generated Video Captions (GVC), where new captions are extracted from the appearance of a video~\cite{DBLP:conf/cvpr/WuLFWO23,DBLP:conf/sigir/DongFZYYG24,DBLP:conf/iclr/XueS0FSLL23}, Generated Multimodality Captions (GMC), where new captions are generated from more video modalities~\cite{DBLP:conf/eccv/WangLLYHCPZWSJLXZHQWW24,DBLP:conf/nips/ChenLWZSZL23} and Questions-Answers (Q$\&$A) systems~\cite{DBLP:conf/iccv/LiangA23,DBLP:conf/emnlp/HanPLLK24}.

\textbf{Generated Video Captions (GVC)} are new captions generated from videos, enabling the creation of a set of different textual descriptions for the same video content. These captions can differ in content, describing different aspects of the same video, or in style, using different textual styles to describe the video. Therefore, having so many new captions can introduce relevant auxiliary input data that can facilitate the alignment between modalities. 

Cap4Video~\cite{DBLP:conf/cvpr/WuLFWO23} integrates auxiliary captions into video retrieval at three levels. First, the generated captions, paired with their corresponding videos, create additional positive training samples. Then, more discriminative video representations are obtained by modelling the interaction between videos and their generated captions. Finally, video-caption similarity scores are incorporated with query-video scores to boost final retrieval performance.
In~\cite{DBLP:conf/sigir/DongFZYYG24}, a key-frame-based approach generates concise and accurate auxiliary captions from visually significant frames. These captions are used to refine the original captions via LLMs and then combined with key frames during training. Similarly, CLIP-ViP~\cite{DBLP:conf/iclr/XueS0FSLL23} generates new captions for the middle frame of each video using captioner models and introduces video proxy tokens that enable CLIP to capture temporal information across frames through a proxy-guided attention mechanism.

Both In-Style~\cite{DBLP:conf/iccv/ShvetsovaKSK23} and VIIT~\cite{DBLP:conf/cvpr/00060ZWCMSA0GKY24} generate high-quality pseudo-captions from uncurated web videos. While In-Style emphasises stylistic alignment by training a captioner to emulate the style of text queries, VIIT introduces a two-stage fine-tuning strategy that alternates tuning the visual encoder and LLM, explicitly targeting temporal and causal reasoning. Finally, T2VIndexer~\cite{DBLP:conf/mm/Li0GL0024} generates auxiliary captions using VLMs and proposes a novel caption-based indexing approach. By clustering videos based on these captions, the model maps text queries directly to relevant clusters, significantly reducing the retrieval time while maintaining high performance.

LaViLa~\cite{DBLP:conf/cvpr/0006MKG23} generates detailed and diverse captions for egocentric videos using two distinct types of LLM-based supervision. A visual-conditioned LLM, i.e. NARRATOR, generates captions for both existing and newly sampled clips, whereas the REPHRASER is a standard LLM that paraphrases existing clip narrations, effectively augmenting the dataset with diverse textual variations.
Building on the stylistic alignment idea from~\cite{DBLP:conf/iccv/ShvetsovaKSK23}, EgoInstructor~\cite{DBLP:conf/cvpr/XuHHCZFX24} proposes an egocentric video retrieval method that integrates third-person instructional captions. Given that egocentric videos are limited to the viewpoint of the camera wearer, this approach enhances video-text representations through cross-view retrieval, bridging egocentric and exocentric perspectives.

\textbf{Generated Multimodal Captions (GMC)} are captions generated from additional modalities, such as audio and speech, that are then fused via LLMs to obtain a multimodal caption. Even though these works~\cite{DBLP:conf/nips/ChenLWZSZL23,DBLP:conf/eccv/WangLLYHCPZWSJLXZHQWW24} incorporate these additional modalities in their architecture as auxiliary information, a key innovation lies in using them to generate omni-modality captions.

VAST~\cite{DBLP:conf/nips/ChenLWZSZL23} generates omni-modality captions to train a video-language model that aligns inputs from vision, audio, and subtitles. Caption generation follows a two-stage process: first, separate vision and audio captioners produce modality-specific captions; second, an LLM fuses these captions with subtitle information to create a single multimodal textual description. Its architecture consists of dedicated modality-specific encoders and a text encoder, trained jointly using contrastive, matching, and generative losses. Similarly, InternVideo2~\cite{DBLP:conf/eccv/WangLLYHCPZWSJLXZHQWW24} generates individual captions from video, audio, and speech inputs, which are then fused through an LLM into a unified description. The model is trained in three stages. First, it learns spatio-temporal features by reconstructing masked video tokens with expert supervision; then it aligns cross-modal features via contrastive learning using fused captions; finally, it connects the video encoder to a large LLM for next-token prediction, enabling open-ended task performance.

\textbf{Questions-Answers (Q\&A)} consists of chat-based interfaces and iterative Q$\&$A mechanisms integrated into video retrieval models to enhance alignment between user queries and complex video content. These systems refine the interaction between video and text embeddings iteratively, achieving higher retrieval performance.

IVR-QA~\cite{DBLP:conf/iccv/LiangA23} generates questions about video content (objects, actions, scenes) to simulate user interactions. A captioner first produces captions for a gallery of videos, which are then processed by a Question Generator module to create a query-specific question. The Answer Generator module produces an answer, based on the video target, that is added to the original textual query to create a new query. Two question-generation strategies are explored: a heuristic approach, which extracts keywords from the original query to generate questions, and a parametric approach, which generates questions based on the query and the top-n retrieved videos for the current query.
Similarly, MERLIN~\cite{DBLP:conf/emnlp/HanPLLK24} introduces a training-free pipeline using LLMs to generate questions from captions and simulate user responses based on target videos, adjusting dynamically the embeddings to capture user intent. 

\section{Auxiliary Information from Text}\label{sec4}

This section reviews works that incorporate additional text-extracted information, emphasising how it is integrated into the proposed methods. Methods using textual info are reviewed in Sec.~\ref{subsec4:textual}, while Sec.~\ref{subsec4:text_generated} covers methods where generative models generate extra data. Tab.~\ref{tab:explanation_auxiliary_text} shows the categorisation of this auxiliary information by also giving a general definition for each one.

\begin{table*}[t!]
\centering
\resizebox{\linewidth}{!}{%
\begin{tabular}{llll}

& &Class
& Auxiliary Information \\
\hline

\multirow{6}{*}{\centering Textual Info}&\multirow{3}{*}{\centering Sentence-Level} & Temporal Dependency (TD)
 & Temporal information derived from captions.\\
 
 && Text Captions (TC)
 & Additional captions available in benchmark datasets.\\
&& Text Experts (TE)
 & Information extracted by specialised pre-trained models.
 \\\cline{2-4}

&\multirow{3}{*}{\centering Word-Level}
 & Part of Speech (POS)
 & Linguistic categories extracted from the caption.\\

 && POS Captions (POS-C)
 & Original captions altered using POS. \\
 && Hierarchy Dependency (HD)
 & Semantic or syntactic POS relationships. \\\hline

\multirow{2}{*}{\centering Generated Info}&&Generated Text Captions (GTC)& Additional captions generated from the original.\\

&&Languages (L)& Translations of captions into other languages.\\\hline\hline
\end{tabular}}
\caption{Explanation of different text-extracted auxiliary information. This information is divided into two main groups: Textual and Generated Info. The Textual Info group is furthermore divided into Sentence-Level and Word-Level subgroups.}
\label{tab:explanation_auxiliary_text}

\end{table*}

\subsection{Textual Info}\label{subsec4:textual}
Captions provide a textual description of the content shown in a video. However, videos are more complex and possess a higher information density compared to text, as textual descriptions often focus on specific segments (frames) or elements (ROIs) rather than covering the full range of events or visual details occurring during the video~\cite{DBLP:conf/cvpr/GortiVMGVGY22}. For instance, given a video of someone baking a cake, the corresponding caption might be ``A person mixes butter and bakes a cake''. This caption only captures a few elements of the video, such as the step of mixing, without fully describing every step or other details shown in the video, such as the kitchen and the tools used.
However, many studies have tried to integrate auxiliary information extracted from the text modality into retrieval methods.
These works can be divided into two main groups: Sentence-Level and Word-Level. Sentence-Level methods rely on auxiliary information extracted globally from the text, whereas Word-Level methods involve local information derived from the individual words that compose the sentence.

\subsubsection{Sentence-Level}
Sentence-Level methods use different textual information including Temporal Dependency (TD)~\cite{DBLP:conf/cvpr/MiechASLSZ20}, which leverage neighbouring captions, Text Captions (TC)~\cite{DBLP:conf/iclr/Patrick0AMHHV21,DBLP:conf/eccv/FangWZHH22,DBLP:journals/tmm/WangGYX21,DBLP:conf/eccv/WangWNR22}, which involve additional captions present in the datasets, and Text Experts (TE)~\cite{DBLP:journals/tmm/0001ZXJ021,DBLP:conf/iccv/CroitoruBLJZAL21}, which use pre-trained models to extract different textual features from the original captions.

\textbf{Temporal Dependency (TD)} consider temporal information of captions over time. MIL-NCE~\cite{DBLP:conf/cvpr/MiechASLSZ20} addresses the challenge of temporal misalignment between clips and their captions in instructional videos by allowing multiple captions to be used as potential positive matches for a given clip. To achieve this, MIL-NCE introduces a modified NCE loss, where a temporal window of neighbouring captions around each clip is considered as positive candidates. This strategy helps mitigate video-text misalignment, enabling the model to learn a shared embedding space in which videos and captions are better aligned.

\textbf{Text Captions (TC)} refers to multiple human-labelled captions for each video in benchmark datasets. Few works have proposed using these available captions to obtain different descriptions of the same video without generating this information via generative models.
Among these, Support-Set~\cite{DBLP:conf/iclr/Patrick0AMHHV21} reconstructs each caption as a weighted combination of textual descriptions from a support set of captions from semantically similar videos. In this way, the model learns representations that generalise more effectively across variations in phrasing and content.
The LINAS framework~\cite{DBLP:conf/eccv/FangWZHH22} instead defines a support set of captions associated with the same video. Then, a teacher model aggregates the available captions through an attention mechanism that assigns weights according to similarity with the query, and a student model emulates these enhanced embeddings during inference, where extra captions are unavailable. 

Distinct from LINAS, MQVR~\cite{DBLP:conf/eccv/WangWNR22}  integrates multiple captions both during training and inference. It proposes several aggregation strategies, including similarity-based weighting and direct feature fusion, allowing additional captions to contribute to the retrieval process at test time. 
In contrast, CF-GNN~\cite{DBLP:journals/tmm/WangGYX21} introduces a Coarse-to-Fine Graph Neural Network that leverages additional captions as nodes in a graph structure, which captures relationships between the video, its primary caption, and the auxiliary textual descriptions. These extra captions enrich the graph with diverse textual perspectives of video content and refine alignment through multi-step reasoning. The coarse-to-fine strategy progressively removes nodes, keeping only the most discriminative ones.

\textbf{Text Experts (TE)} involves extracting different coarse text features from various pre-trained architectures to obtain separate representations of the same caption and then fusing them for the final caption embedding. In~\cite{DBLP:journals/tmm/0001ZXJ021}, the Sentence Encoder Assembly (SEA) framework leverages multiple sentence encoders to capture diverse text representations. SEA uses a multi-loss learning framework to map queries to distinct spaces, optimised with dedicated loss functions, that are then combined to generate a unified embedding space.
Similarly, TeachText~\cite{DBLP:conf/iccv/CroitoruBLJZAL21} proposes a teacher-student distillation framework, where multiple teacher models, each based on a different pre-trained text encoder, generate diverse text representations. During training, the outputs of the teachers are integrated into a unified similarity matrix between text and video features. Then, a student model is trained to approximate this matrix using only a single text encoder at inference.

\subsubsection{Word-Level}
Different from Sentence-level methods, these works involve local extra information by leveraging the semantics and linguistic properties of words in a sentence. Some methods directly extract and use Part-of-Speech (POS) tags~\cite{DBLP:conf/mm/HanCXZZC21,DBLP:conf/cvpr/YuKCK17,DBLP:conf/iccv/WrayCLD19,DBLP:journals/corr/abs-2109-04290,DBLP:conf/aaai/XuXCC15,DBLP:conf/iccv/YangBG21,DBLP:conf/cvpr/GeGLLSQL22,DBLP:conf/icassp/LeK0L24}. Other works involve POS tags to generate new captions, i.e. POS Captions (POS-C)~\cite{DBLP:conf/mm/WangCH022}. Finally, several approaches utilise the relations between words or POS tags, i.e. Hierarchy Dependency (HD)~\cite{DBLP:conf/sigir/YangD0W0C20,DBLP:journals/mms/LvSN24,DBLP:conf/aaai/XuXCC15,DBLP:conf/cvpr/ChenZJW20,DBLP:conf/mm/WuHTLL21,DBLP:journals/corr/abs-2404-14066}.

\textbf{Part-of-Speech (POS)} tags refer to word categories based on their grammatical roles (e.g. nouns, verbs, adjectives, and adverbs) that help identify the function of each word in a sentence.
Some of these works use this information to learn specific embedding spaces~\cite{DBLP:conf/iccv/WrayCLD19,DBLP:journals/pami/DongLXYYWW22}. More precisely, JPoSE~\cite{DBLP:conf/iccv/WrayCLD19} disentangles captions into separate verb and noun embeddings that are then integrated into a unified space, and Dual Encoding~\cite{DBLP:journals/pami/DongLXYYWW22} leverages POS-tagged captions for a hybrid concept-based representation. In contrast, other works align POS tags with corresponding video features in a dual-stream architecture~\cite{DBLP:journals/corr/abs-2109-04290}, model relationships between textual phrases and frame-level features via graph auto-encoder models~\cite{DBLP:conf/mm/HanCXZZC21}, or define a concept word detector that identifies high-level semantic concepts in videos~\cite{DBLP:conf/cvpr/YuKCK17}.

The use of POS tags has been refined by a few works~\cite{DBLP:conf/iccv/YangBG21,DBLP:conf/cvpr/GeGLLSQL22,DBLP:conf/icassp/LeK0L24} with the introduction of novel training paradigms and auxiliary tasks. More precisely, a token-aware contrastive learning strategy~\cite{DBLP:conf/iccv/YangBG21} and a pre-text task involving noun- and verb-based multiple-choice questions~\cite{DBLP:conf/cvpr/GeGLLSQL22} are proposed to improve token-level video-text alignment. In contrast, WAVER~\cite{DBLP:conf/icassp/LeK0L24} proposes to use POS tags within a cross-domain knowledge distillation framework. A Video Content Dictionary (VCD) is constructed by extracting verb phrases from each caption fed into a VLM (i.e. CLIP) through manually designed prompts to select the most relevant phrases for all the training videos. This VCD acts as an open-vocabulary knowledge source derived from the pre-trained CLIP that is then transferred into a vision-based student model, with the VLM implicitly acting as the teacher to guide downstream video–text alignment.

\textbf{POS Captions (POS-C)} represent additional captions that are generated by modifying the original caption using POS information. CLIP-bnl~\cite{DBLP:conf/mm/WangCH022} addresses the limitation of retrieval models to handle negations in textual queries by constructing two types of queries: negated queries, where a verb in the caption is negated, and composed queries, where positive and negative clauses are combined via template-based generation. During training, a Simple Negation Learning pushes negated captions away from their matching video, treating them as negative examples. In addition, a Bidirectional Negation Learning ensures that negated captions are less similar than the original captions, but not completely unrelated. This dual mechanism helps the model understand what is being negated without losing the rest of the context.

\textbf{Hierarchy Dependency (HD)} improves retrieval performance by leveraging the semantic or syntactic information of words or POS.  A few studies explore graph-based methods for modelling dependency between words~\cite{DBLP:conf/sigir/YangD0W0C20} or POS~\cite{DBLP:journals/mms/LvSN24,DBLP:conf/mm/WuHTLL21,DBLP:conf/cvpr/ChenZJW20}, where nouns and verbs are considered as entity and action nodes, respectively. 

HGR~\cite{DBLP:conf/cvpr/ChenZJW20} aligns parsed semantic roles with video representations through attention-based graph reasoning. This idea was extended in~\cite{DBLP:conf/mm/WuHTLL21} by simultaneously applying hierarchical parsing to videos and captions.
MT-AGCN~\cite{DBLP:journals/mms/LvSN24} utilises GCNs to enhance textual encoding by modelling semantic relationships between actions and entities extracted via POS.

Different from graph-based methods, SHE-Net~\cite{DBLP:journals/corr/abs-2404-14066} leverages the grammatical dependencies in text via a syntax tree hierarchy that guides the creation of a video syntax hierarchy, aligning verbs with frames and nouns with spatial regions. Similarly, TCE~\cite{DBLP:conf/sigir/YangD0W0C20} introduces a Latent Semantic Tree to decompose a textual query and captures relationships between words in a text.

\subsection{Generated Info}\label{subsec4:text_generated}
Many works focused on generating new auxiliary information from the original caption with the use of generative models. This generated information from text can be divided into Generated Text Captions (GTC)~\cite{DBLP:conf/eccv/ShvetsovaKHRSK24,bai2025bridging, DBLP:conf/iclr/XuWDSZJ25}, where new captions are generated from the original one, and Languages (L)~\cite{DBLP:conf/naacl/HuangPHNMH21,DBLP:journals/ir/MadasuASRTBL23,DBLP:conf/icassp/RouditchenkoCSTFKKHKG23}, where the original English caption is translated into different languages.

\textbf{Generated Text Captions (GTC)} consists of enriched captions generated from the original one, creating context-aware descriptions, and refining captions to better describe a video's visual content. HowToCaption~\cite{DBLP:conf/eccv/ShvetsovaKHRSK24} and GQE~\cite{bai2025bridging} leverage LLMs to generate high-quality, descriptive captions that significantly enhance retrieval performance. Specifically, HowToCaption transforms noisy ASR subtitles into human-like captions, and GQE adopts a query expansion strategy, where multiple queries are generated from the original one via LLMs during testing. A query selection mechanism filters these queries based on their relevance and diversity. Then, the remaining queries are aggregated to create the enhanced query embedding used in the retrieval process

In contrast, EgoNCE++~\cite{DBLP:conf/iclr/XuWDSZJ25} enhances egocentric video retrieval, a domain characterised by detailed hand-object interactions and complex first-person actions, by introducing an asymmetric contrastive loss framework. For the video-to-text matching, EgoNCE++ generates hard negative captions by changing key verbs or nouns using LLMs, forcing the model to learn finer semantic distinctions. For the text-to-video matching, the model learn to recognise nouns by aggregating video features associated with similar nouns.

\textbf{Languages (L)} consists of additional captions obtained by translating the original English caption. This makes the retrieval task useful worldwide and addresses the performance gap between English and other languages. When non-English captions are unavailable, state-of-the-art machine translation methods translate the original English captions. Specifically, MMP~\cite{DBLP:conf/naacl/HuangPHNMH21} uses a multilingual extension of the HowTo100M~\cite{DBLP:conf/naacl/HuangPHNMH21} and employs a transformer-based architecture with intra-modal, inter-modal, and conditional cross-lingual contrastive objectives. On the contrary, MuMUR~\cite{DBLP:journals/ir/MadasuASRTBL23} employs a dual cross-modal encoder built upon CLIP to explicitly align multilingual captions with visual representations. C2KD~\cite{DBLP:conf/icassp/RouditchenkoCSTFKKHKG23} introduces a cross-lingual student–teacher knowledge distillation framework, where multiple teacher models, each trained on English queries, guide a multilingual student model via a cross-entropy-based distillation objective. This approach effectively transfers semantic knowledge from English supervision into non-English retrieval tasks and mitigates the reliance on large-scale human-annotated multilingual datasets.

\section{Performance with Auxiliary Information}\label{sec5}

Sec.~\ref{sec.datasets} details the most common datasets used for Text-to-Video retrieval, emphasising the auxiliary information available, and Sec.~\ref{results} compares the results from the methods described previously.

\subsection{Auxiliary Information in Datasets}\label{sec.datasets}
We categorise the video retrieval datasets, comprising video clips and annotated descriptions, into Pre-training and Downstream Datasets. Pre-training datasets are used solely for pre-training due to their noisy annotations or scale before evaluating the methods on downstream tasks. We have two common Pre-training video retrieval datasets: \textbf{HowTo100M}~\cite{DBLP:conf/iccv/MiechZATLS19} and \textbf{Ego4D}~\cite{DBLP:conf/cvpr/GraumanWBCFGH0L22}. On the contrary, Downstream Datasets directly assess the performance of the method on retrieval tasks. We have 8 common Downstream video retrieval datasets: \textbf{MSR-VTT}~\cite{DBLP:conf/cvpr/XuMYR16}, \textbf{MSVD}~\cite{DBLP:conf/acl/ChenD11}, \textbf{ActivityNet Captions}~\cite{DBLP:conf/iccv/KrishnaHRFN17}, \textbf{DiDeMo}~\cite{DBLP:conf/iccv/HendricksWSSDR17}, \textbf{LSMDC}~\cite{DBLP:conf/cvpr/RohrbachRTS15}, \textbf{YouCook2}~\cite{DBLP:conf/aaai/ZhouXC18}, \textbf{VATEX}~\cite{DBLP:conf/iccv/WangWCLWW19} 
and \textbf{EPIC-KITCHENS-100}~\cite{Damen2018EPICKITCHENS}.
We focus on the auxiliary information provided by video retrieval datasets, while the dataset details are introduced in Sec.~\ref{appendix_datasets} of the Appendix.

We first identify any additional information in each dataset beyond the video-text pairs to conduct this analysis and then categorise this using the auxiliary classes previously defined in Tab.~\ref{tab:explanation_auxiliary_video} (video-extracted information) and Tab.~\ref{tab:explanation_auxiliary_text} (text-extracted information). These auxiliary classes group similar additional information under unified labels, such as Audio, Video Temporal Context (VTC), Textual Dependencies (TD) and Video Experts (VE). The complete list of auxiliary information available in each dataset and its mapping to the corresponding auxiliary classes is presented in Tab.~\ref{tab:auxiliary_per_dataset}.

\begin{table}
\centering
\resizebox{\linewidth}{!}{
\begin{tabular}{l|l|c|c|ccc|ccc|c|c|c|c|c|c}
\multicolumn{2}{c|}{} & \multicolumn{10}{c|}{Video-Extracted}& \multicolumn{4}{c}{Text-Extracted} \\
\cline{3-16}
\multicolumn{2}{c|}{} & \multirow{2}{*}{Audio} & \multirow{2}{*}{VTC} & \multicolumn{3}{c|}{MM} & \multicolumn{3}{c|}{VE} &ROI& VT & \multirow{2}{*}{TD} & \multirow{2}{*}{TC} & \multirow{2}{*}{POS} & \multirow{2}{*}{L} \\\cline{5-12}
 & Dataset & & & Gaze & Stereo Cam. & 3D Scans & Multi Cam. & Faces & Optical Flow &Objects/Hands& Task Class & & & & \\
\hline
\multirow{3}{*}{Pre-training}
& HowTo100M~\cite{DBLP:conf/iccv/MiechZATLS19} & {\color{blue}$\checkmark$} & {\color{blue}$\checkmark$} & $\times$ & $\times$ & $\times$ & $\times$ & $\times$ & $\times$ &$\times$ & {\color{blue}$\checkmark$} & {\color{orange}$\checkmark$} & $\times$ & $\times$ &{\color{orange}$\checkmark$}  \\
& Ego4D~\cite{DBLP:conf/cvpr/GraumanWBCFGH0L22} & {\color{blue}$\checkmark$} & {\color{blue}$\checkmark$} & {\color{blue}$\checkmark$} & {\color{blue}$\checkmark$} & {\color{blue}$\checkmark$} & {\color{blue}$\checkmark$} & {\color{blue}$\checkmark$} & $\times$ &{\color{blue}$\checkmark$} & $\times$ & {\color{orange}$\checkmark$} & {\color{orange}$\checkmark$}  & {\color{orange}$\checkmark$} & $\times$\\
\hline
\multirow{10}{*}{Downstream}
& MSR-VTT~\cite{DBLP:conf/cvpr/XuMYR16} & {\color{blue}$\checkmark$} & $\times$ & $\times$ & $\times$ & $\times$ & $\times$ & $\times$ & $\times$ & $\times$ & $\times$ &$\times$ & {\color{orange}$\checkmark$} & $\times$ & {\color{orange}$\checkmark$} \\
& MSVD~\cite{DBLP:conf/acl/ChenD11} & $\times$ & $\times$ & $\times$ & $\times$ & $\times$ & $\times$ & $\times$ & $\times$ & $\times$ & $\times$ & $\times$ &{\color{orange}$\checkmark$} & $\times$ & {\color{orange}$\checkmark$} \\
& ActivityNet C.~\cite{DBLP:conf/iccv/KrishnaHRFN17} & {\color{blue}$\checkmark$} & {\color{blue}$\checkmark$} & $\times$ & $\times$ & $\times$ & $\times$ & $\times$ & $\times$ & $\times$ & $\times$ &{\color{orange}$\checkmark$} & $\times$ & $\times$ & $\times$ \\
& DiDeMo~\cite{DBLP:conf/iccv/HendricksWSSDR17} & $\times$ & {\color{blue}$\checkmark$} & $\times$ & $\times$ & $\times$ & $\times$ & $\times$ & $\times$ & $\times$ & $\times$ &{\color{orange}$\checkmark$} & $\times$ & $\times$ & $\times$ \\
& LSMDC~\cite{DBLP:conf/cvpr/RohrbachRTS15} & {\color{blue}$\checkmark$} & $\times$ & $\times$ & $\times$ & $\times$ & $\times$ & $\times$ & $\times$ & $\times$ & $\times$ & $\times$ &{\color{orange}$\checkmark$}& $\times$ & $\times$ \\
& YouCook2~\cite{DBLP:conf/aaai/ZhouXC18} & {\color{blue}$\checkmark$} & {\color{blue}$\checkmark$} & $\times$ & $\times$ & $\times$ & $\times$ & $\times$ & $\times$ & $\times$ & $\times$& {\color{orange}$\checkmark$} & $\times$ & $\times$ &{\color{orange}$\checkmark$}\\
& VATEX~\cite{DBLP:conf/iccv/WangWCLWW19} & {\color{blue}$\checkmark$} & $\times$ & $\times$ & $\times$ & $\times$ & $\times$ & $\times$ & $\times$ & $\times$ & $\times$ &$\times$ & {\color{orange}$\checkmark$} & $\times$ & {\color{orange}$\checkmark$} \\
& EPIC-K~\cite{Damen2018EPICKITCHENS} & {\color{blue}$\checkmark$} & {\color{blue}$\checkmark$} & $\times$ & $\times$ & $\times$ & $\times$ & $\times$ & {\color{blue}$\checkmark$} & $\times$ & $\times$& {\color{orange}$\checkmark$} & $\times$ & {\color{orange}$\checkmark$} & {\color{orange}$\checkmark$} \\
\hline
\end{tabular}
}
\caption{Auxiliary information, grouped by auxiliary classes, available in both Pre-training and Downstream datasets. {\color{blue}$\checkmark$} and {\color{orange}$\checkmark$} indicate the video- and text-extracted information available, respectively. $\times$ indicates the missing auxiliary information. ActivityNet C. denotes ActivityNet Captions, and EPIC-K refers to EPIC-KITCHENS-100. The class labels are defined in Tab.~\ref{tab:explanation_auxiliary_video} (video-extracted) and Tab.~\ref{tab:explanation_auxiliary_text} (text-extracted), respectively.}
\label{tab:auxiliary_per_dataset}
\end{table}

Among Pre-training datasets, Ego4D~\cite{DBLP:conf/cvpr/GraumanWBCFGH0L22} provides $11$ types of auxiliary information, $8$ video-extracted: Audio, Video Temporal Context (VTC)~\cite{DBLP:conf/nips/LinWSWYXGTZKCWD22}, Gaze (MM), Stereo Cameras (MM), 3D Scans (MM), Multi Cameras (VE), Faces (VE). and Objects/Hands annotations (ROI); and $3$ text-extracted: Temporal Dependency (TD), Text Captions (TC) and human-labelled Part-of-Speech (POS) tags. HowTo100M~\cite{DBLP:conf/iccv/MiechZATLS19} offers a limited set of $3$ video-extracted information: Audio, Video Temporal Context (VTC) and task classes (VT); and  $2$ text-extracted: Temporal Dependency (TD) and multiple Languages (L)~\cite{DBLP:conf/naacl/HuangPHNMH21}.

On the contrary, Downstream datasets provide a limited amount of auxiliary information compared to the Pre-training datasets. An exception is EPIC-KITCHENS-100~\cite{Damen2018EPICKITCHENS}, which includes $3$ types of video-extracted information: Audio, Video Temporal Context (VTC) and Optical Flow (VE); and $3$ types of text-extracted information: Temporal Dependency (TD), human-labelled Part-of-Speech (POS) tags and multiple Languages (L). However, some resources, such as multiple languages (i.e. English, Italian, Greek, Chinese), have never been explored in detail in this dataset. Other downstream datasets usually offer fewer types of auxiliary information that are heavily biased towards text-extracted information, i.e. Temporal Dependency (TD), Text Captions (TC) and multiple Languages (L), while offering only Audio and Video Temporal Context (VTC) as video-extracted information.

The growing availability of multilingual datasets is primarily driven by dataset expansions rather than human-labelled annotations. While VATEX~\cite{DBLP:conf/iccv/WangWCLWW19} provides English and Chinese languages others, such as Multi-HowTo100M~\cite{DBLP:conf/naacl/HuangPHNMH21}, Multi-MSRVTT~\cite{DBLP:conf/naacl/HuangPHNMH21}, Multi-YouCook2~\cite{DBLP:conf/icassp/RouditchenkoCSTFKKHKG23} and different versions of MSVD~\cite{DBLP:journals/mt/CitamakCKEEMS21,DBLP:journals/corr/abs-2306-11341} rely on automatic machine translation.
Even though this strategy significantly reduces the cost of building multilingual datasets, it introduces linguistic inconsistencies compared to human-curated captions.

\paragraph{Conclusions and Findings}

Pre-training datasets such as Ego4D and HowTo100M are richer in terms of quantity and diversity, showcasing a prevalence of video-extracted information. In contrast, Downstream datasets have fewer types and focus more on text-extracted, commonly providing multiple captions and/or multilingual translations. However, many multilingual datasets are generated by involving machine translation models, which may introduce inconsistencies that can affect retrieval performance.

Moreover, while auxiliary classes such as Audio, Video Temporal Context (VTC), and Textual Context (TD) are commonly available, specialised information such as Multi-Modalities (MM), human-labelled Regions of Interest (ROI), and hand-refined POS Taxonomies (POS) remain rare. Their scarcity reflects the challenges associated with collecting these annotations at scale.

In conclusion, we argue that understanding the distribution and limitations of available auxiliary information in video retrieval datasets is essential for guiding future research directions and design new video retrieval datasets, which incorporate additional richer and diverse modalities and human-labelled information.

\subsection{Retrieval Results with Auxiliary Information}
\label{results}

We consider all the Downstream datasets to ensure a comprehensive evaluation across general exocentric benchmarks and more specialised egocentric and domain-specific datasets. This enables a robust comparison of retrieval methods under varying visual domains and annotation granularities.

\subsubsection{Evaluation Details.}

We focus only on the text-to-video retrieval task, which remains the most commonly studied retrieval task. Results are reported using standard evaluation metrics: \textbf{Recall@$1$ ($R@1$)} measures the proportion of relevant videos retrieved within the top-ranked result. Higher values indicate better performance. \textbf{Normalised Discounted Cumulative Gain ($nDCG$)} assesses ranking quality by considering both the relevance of retrieved videos and their positions within the ranked list. Higher scores reflect better performance. \textbf{Mean Average Precision ($mAP$)} quantifies the mean average precision across all relevant videos for a set of queries. Higher is again better. 

Given the extensive number of methods discussed, we prioritise these three metrics to improve consistency and readability, omitting other commonly used metrics such as Recall@$5$ ($R@5$), Recall@$10$ ($R@10$) and Median Rank ($MR$), see Sec.~\ref{full_table} in the Appendix for full comparison tables.

A few works do not report the full set of evaluation metrics considered in this comparison. In particular, VATT~\cite{DBLP:conf/nips/AkbariYQCCCG21} and LINAS framework~\cite{DBLP:conf/eccv/FangWZHH22} do not report $R@1$ and HierVL~\cite{DBLP:conf/cvpr/AshutoshGTG23} and EgoInstructor~\cite{DBLP:conf/cvpr/XuHHCZFX24} report only the average $mAP$ and $nDCG$ between text-to-video and video-to-text tasks. We exclude these works from the quantitative comparison tables to maintain consistency and focus exclusively on the text-to-video retrieval. However, their methods were discussed previously to provide a better review of the use of auxiliary information in cross-modal video retrieval.

The evaluation framework consists of two scenarios. The \emph{Zero-Shot} setting, where models are pre-trained on large-scale datasets and evaluated on the test set of Downstream datasets without further fine-tuning, and the \emph{Fine-Tuned} setting, where models are further trained on the specific training set of Downstream datasets, starting from scratch or leveraging pre-trained weights from the large-scale corpus. This dual setting allows a fairer comparison between different approaches.

We first analyse results within single datasets (\emph{Intra-Dataset Analysis}) to identify outperforming methods and auxiliary classes. Then we extend the comparison between multiple datasets (\emph{Overall Analysis}) to identify cross-dataset trends. For clarity, we define methods that leverage video-extracted information as \emph{video-information} and methods that leverage text-extracted information as \emph{text-information}.

\begin{figure}
	\centering
	\includegraphics[width=0.9\linewidth]{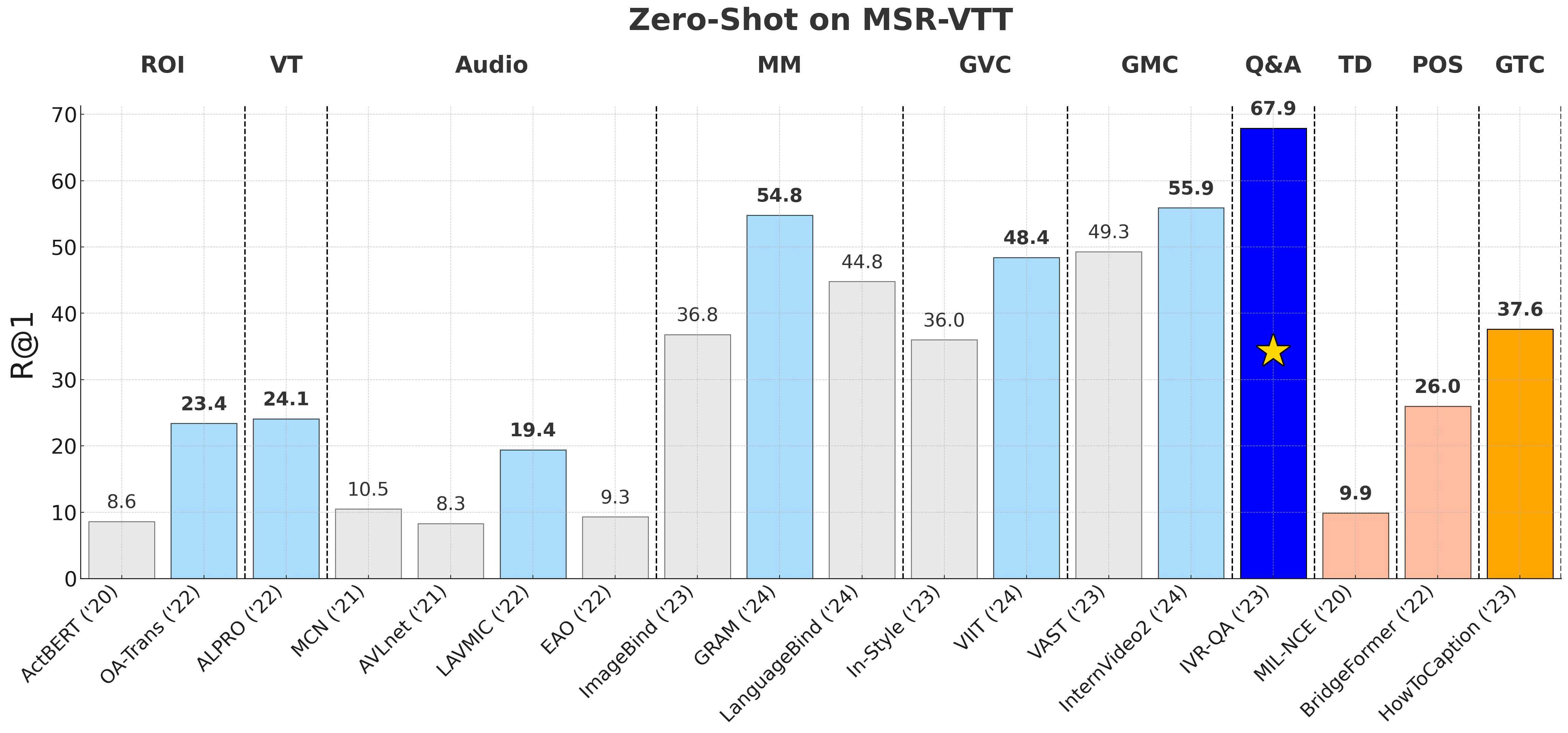}
	\caption{Performance comparison of Zero-Shot models on MSR-VTT across different types of auxiliary information. The \textcolor{gold}{\textbf{star}} absolute best-performing. \textcolor{blue}{\textbf{Blue}} and \textcolor{orange}{\textbf{orange}} bars best-performing across video- and text-extracted classes, respectively. \textcolor{lightskyblue}{\textbf{Light blue}} and \textcolor{lightsalmon}{\textbf{light salmon}} bars best-performing within video- and text-extracted classes, respectively. \textcolor{lightgray}{\textbf{Light grey}} other methods. The class labels are defined in Tab.~\ref{tab:explanation_auxiliary_video} (video-extracted) and Tab.~\ref{tab:explanation_auxiliary_text} (text-extracted).}
	\label{fig:msrvtt_zs_comparison}
\end{figure}

\subsubsection{Intra-Dataset Analysis.}

We begin our comparison by analysing each dataset individually to identify which methods and auxiliary classes achieve the highest performance in both Zero-Shot and Fine-Tuned settings. We report the class of auxiliary information leveraged by each method in parentheses, e.g. IVR-QA (Q$\&$A), for clear comparison.
\begin{figure}
	\centering
	\includegraphics[width=0.9\linewidth]{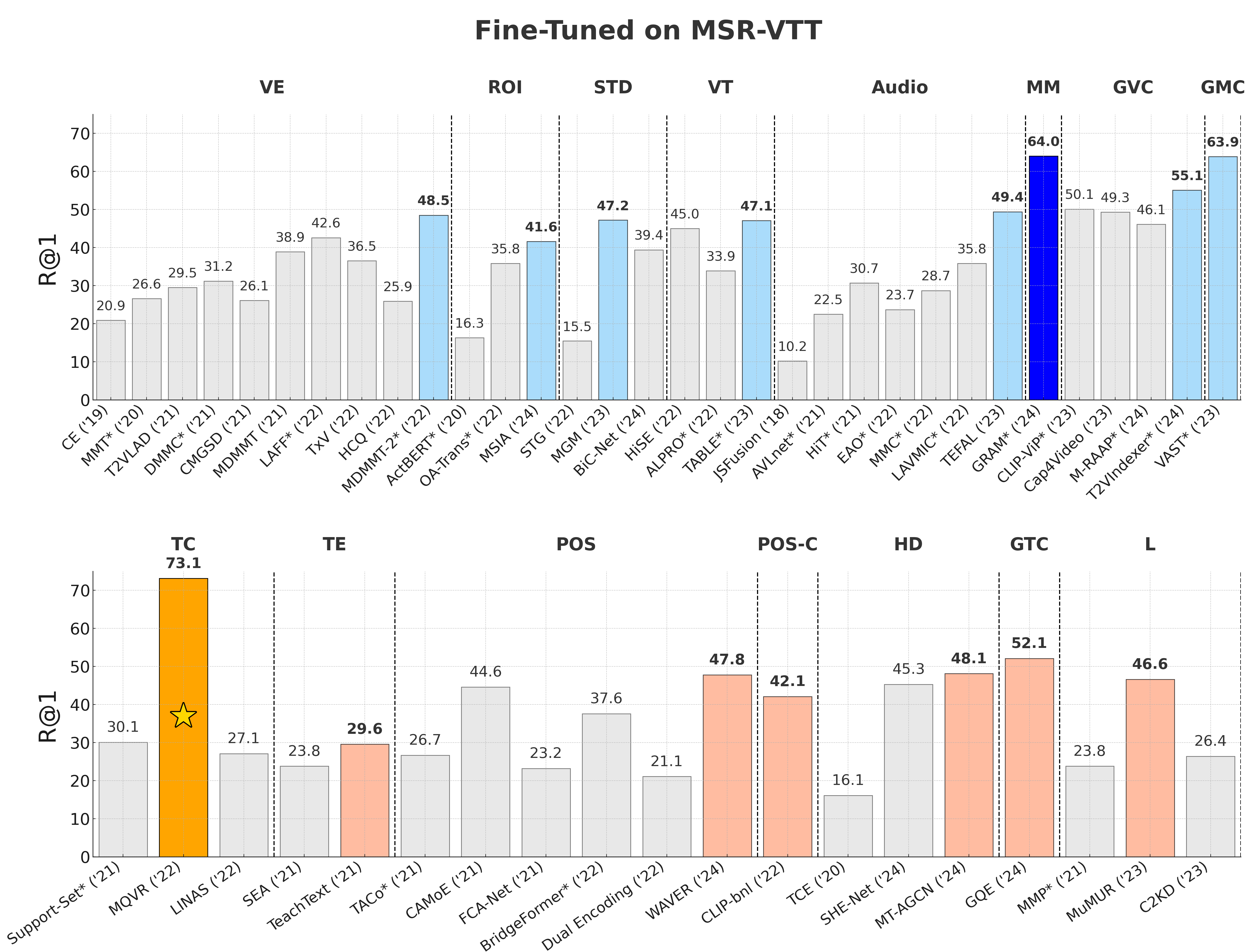}
	\caption{Performance comparison of Fine-Tuned models on MSR-VTT across different types of auxiliary information. The \textcolor{gold}{\textbf{star}} absolute best-performing. \textcolor{blue}{\textbf{Blue}} and \textcolor{orange}{\textbf{orange}} bars best-performing across video- and text-extracted classes, respectively. \textcolor{lightskyblue}{\textbf{Light blue}} and \textcolor{lightsalmon}{\textbf{light salmon}} bars best-performing within video- and text-extracted classes, respectively. \textcolor{lightgray}{\textbf{Light grey}} other methods. $^*$: Pre-training. The class labels are defined in Tab.~\ref{tab:explanation_auxiliary_video} (video-extracted) and Tab.~\ref{tab:explanation_auxiliary_text} (text-extracted).}
\label{fig:msrvtt_ft_comparison}
\end{figure}

\paragraph{Results on MSR-VTT} Fig.~\ref{fig:msrvtt_zs_comparison} and Fig.~\ref{fig:msrvtt_ft_comparison} show the MSR-VTT results in the Zero-Shot and Fine-Tuned settings. In the Zero-Shot setting, video-information methods such as InternVideo2~\cite{DBLP:conf/eccv/WangLLYHCPZWSJLXZHQWW24} (GMC), VIIT~\cite{DBLP:conf/cvpr/00060ZWCMSA0GKY24} (GVC) and GRAM~\cite{DBLP:conf/iclr/CicchettiGSC25} (MM) achieve the best performance in their corresponding classes, with competitive $R@1$ scores of $55.9$, $48.4$ and $54.8$, respectively. However, they are surpassed by the IVR-QA~\cite{DBLP:conf/iccv/LiangA23} (Q$\&$A) with the highest $R@1$ of $67.9$. All these methods outperform the best text-information method HowToCaption~\cite{DBLP:conf/eccv/ShvetsovaKHRSK24} (GTC), which performs with $R@1=37.6$, by a large margin. In the Fine-Tuned setting, strong video-information methods such as GRAM (MM), T2VIndexer~\cite{DBLP:conf/mm/Li0GL0024} (GVC) and VAST~\cite{DBLP:conf/nips/ChenLWZSZL23} (GMC) are outperformed by the video-information MQVR~\cite{DBLP:conf/eccv/WangWNR22} (TC), which achieves the best $R@1$ score of $73.1$.

\begin{figure}
	\centering
	\includegraphics[width=0.9\linewidth]{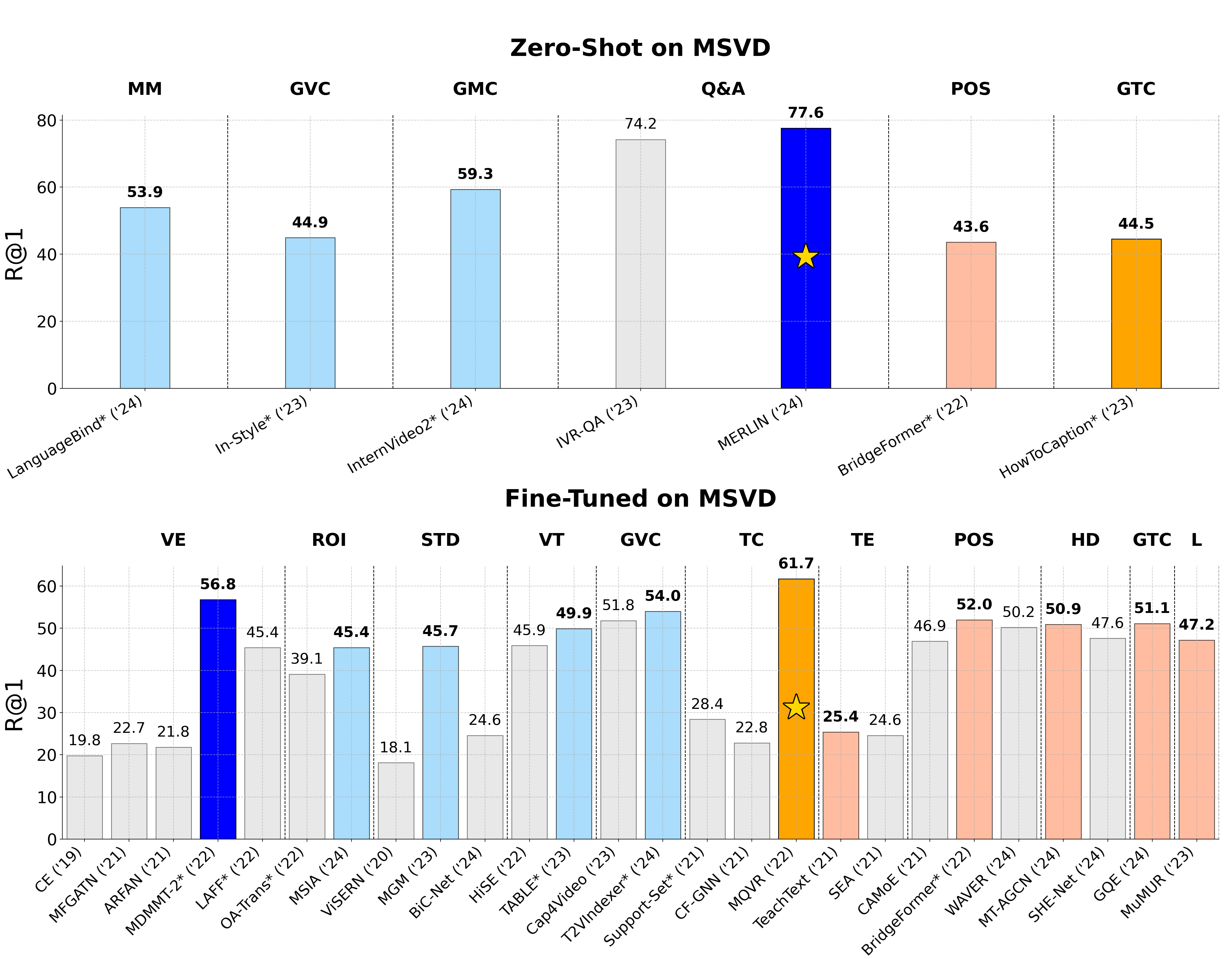}
	\caption{Performance comparison of Zero-Shot/Fine-Tuned models on MSVD across different types of auxiliary information. The \textcolor{gold}{\textbf{star}} absolute best-performing. \textcolor{blue}{\textbf{Blue}} and \textcolor{orange}{\textbf{orange}} bars best-performing across video- and text-extracted classes, respectively. \textcolor{lightskyblue}{\textbf{Light blue}} and \textcolor{lightsalmon}{\textbf{light salmon}} bars best-performing within video- and text-extracted classes, respectively. \textcolor{lightgray}{\textbf{Light grey}} other methods. $^*$: Pre-training. The class labels are defined in Tab.~\ref{tab:explanation_auxiliary_video} (video-extracted) and Tab.~\ref{tab:explanation_auxiliary_text} (text-extracted).}
	\label{fig:msvd_comparison}
\end{figure}
\paragraph{Results on MSVD}
In Fig.~\ref{fig:msvd_comparison}, the results on MSVD reveal a similar conclusion to MSR-VTT. MERLIN~\cite{DBLP:conf/emnlp/HanPLLK24} (Q$\&$A) achieves the highest $R@1$ score of $77.6$ in the Zero-Shot setting, surpassing other video-information methods such as InternVideo2 (GMC) and LanguageBind~\cite{DBLP:conf/iclr/ZhuLNYCWPJZLZ0024} (MM).
Both MERLIN (Q$\&$A) and IVR-QA (Q$\&$A) outperform text-information methods such as HowToCaption (GTC). In the Fine-Tuned setting, MQVR (TC) again outperforms video-information methods, such as MDMMT-2~\cite{DBLP:journals/corr/abs-2203-07086} (VE) and T2VIndexer (GVC), with $R@1=61.7$, confirming the additional value of available text-extracted information (i.e. multiple captions) for retrieval.

\begin{figure}[t]
    \centering
    \begin{minipage}[t]{0.48\linewidth}
        \centering
        \includegraphics[width=\linewidth]{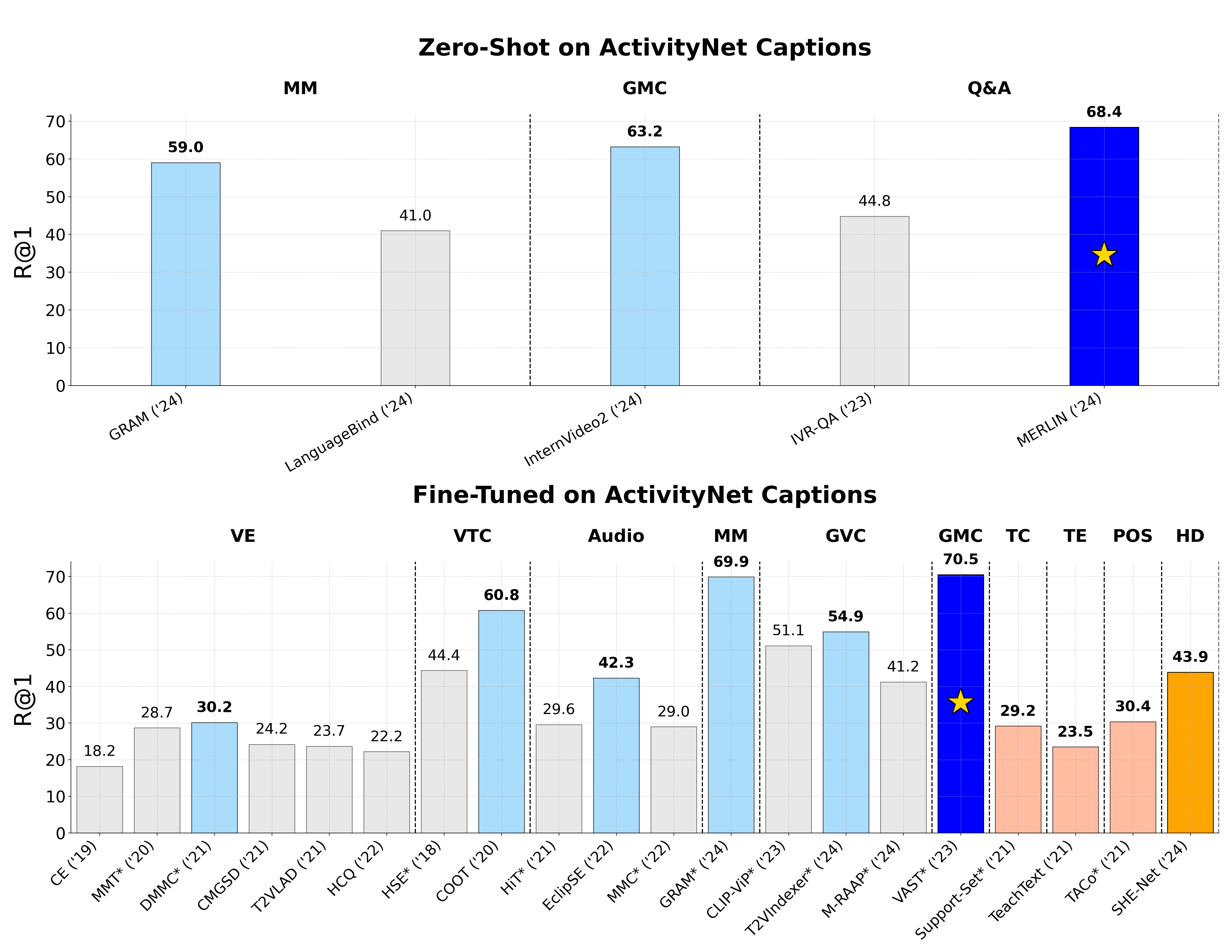}
    \end{minipage}
    \hfill
    \begin{minipage}[t]{0.48\linewidth}
        \centering
        \includegraphics[width=\linewidth]{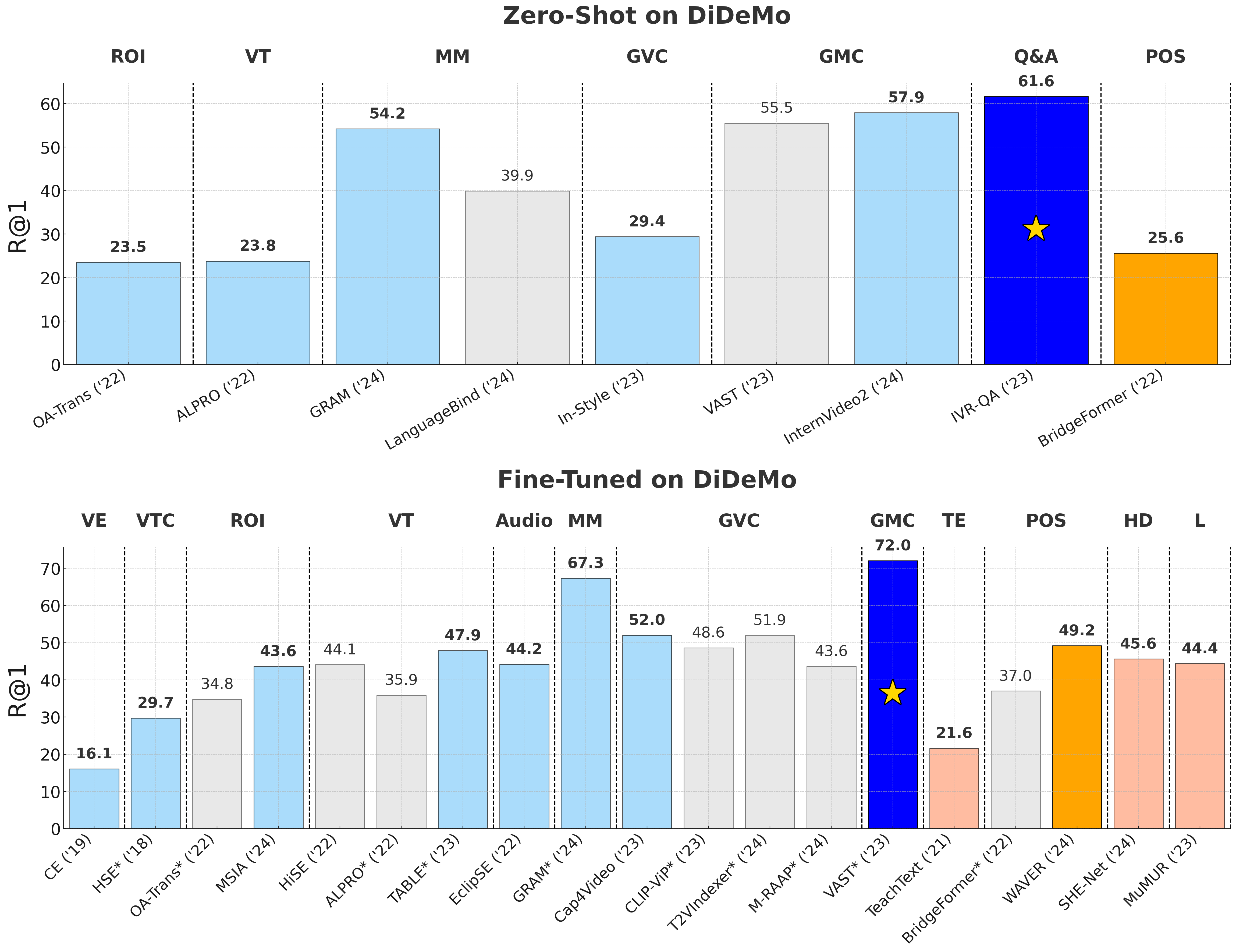}
    \end{minipage}
    	\caption{Performance comparison of Zero-Shot/Fine-Tuned models on ActivityNet Captions (left) and DiDeMo (right) across different types of auxiliary information. The \textcolor{gold}{\textbf{star}} absolute best-performing. \textcolor{blue}{\textbf{Blue}} and \textcolor{orange}{\textbf{orange}} bars best-performing across video- and text-extracted classes, respectively. \textcolor{lightskyblue}{\textbf{Light blue}} and \textcolor{lightsalmon}{\textbf{light salmon}} bars best-performing within video- and text-extracted classes, respectively. \textcolor{lightgray}{\textbf{Light grey}} other methods. $^*$: Pre-training. The class labels are defined in Tab.~\ref{tab:explanation_auxiliary_video} (video-extracted) and Tab.~\ref{tab:explanation_auxiliary_text} (text-extracted).}
        \label{fig:act_didemo_comparison}
\end{figure}

\paragraph{Results on ActivityNet Captions.} Fig.~\ref{fig:act_didemo_comparison} (left) presents the video-paragraph results on ActivityNet Captions. In the Zero-Shot setting, MERLIN outperforms all the other zero-shot methods, including GRAM (MM) and InternVideo2 (GMC), with the highest $R@1$ of $68.4$. In the Fine-Tuned setting, VAST (GMC) achieves the best performance with $R@1=70.5$, surpassing both MM- and GVC-based, and all the text-information methods. Interestingly, COOT~\cite{DBLP:conf/nips/GingZPB20} (VTC), which leverages Video Temporal Context, remains still competitive with $R@1=60.8$, suggesting that temporal modelling continues to provide valuable information in untrimmed videos.

\paragraph{Results on DiDeMo.} 
Fig.~\ref{fig:act_didemo_comparison} (right) shows the video-paragraph results on DiDeMo, which lead to similar conclusions to ActivityNet Captions. IVR-QA (Q$\&$A) achieves the best performance with $R@1=61.6$, outperforming GRAM (MM) and InternVideo2 (GMC), in the Zero-Shot setting. In the Fine-Tuned setting, VAST (GMC) outperforms with $R@1=72.0$ all the other video-information approaches, such as GRAM (MM) and Cap4Video~\cite{DBLP:conf/cvpr/WuLFWO23} (GVC) and text-information methods,such as WAVER~\cite{DBLP:conf/icassp/LeK0L24} (POS).

\begin{figure}[t]
    \centering
    \begin{minipage}[t]{0.48\linewidth}
        \centering
        \includegraphics[width=\linewidth]{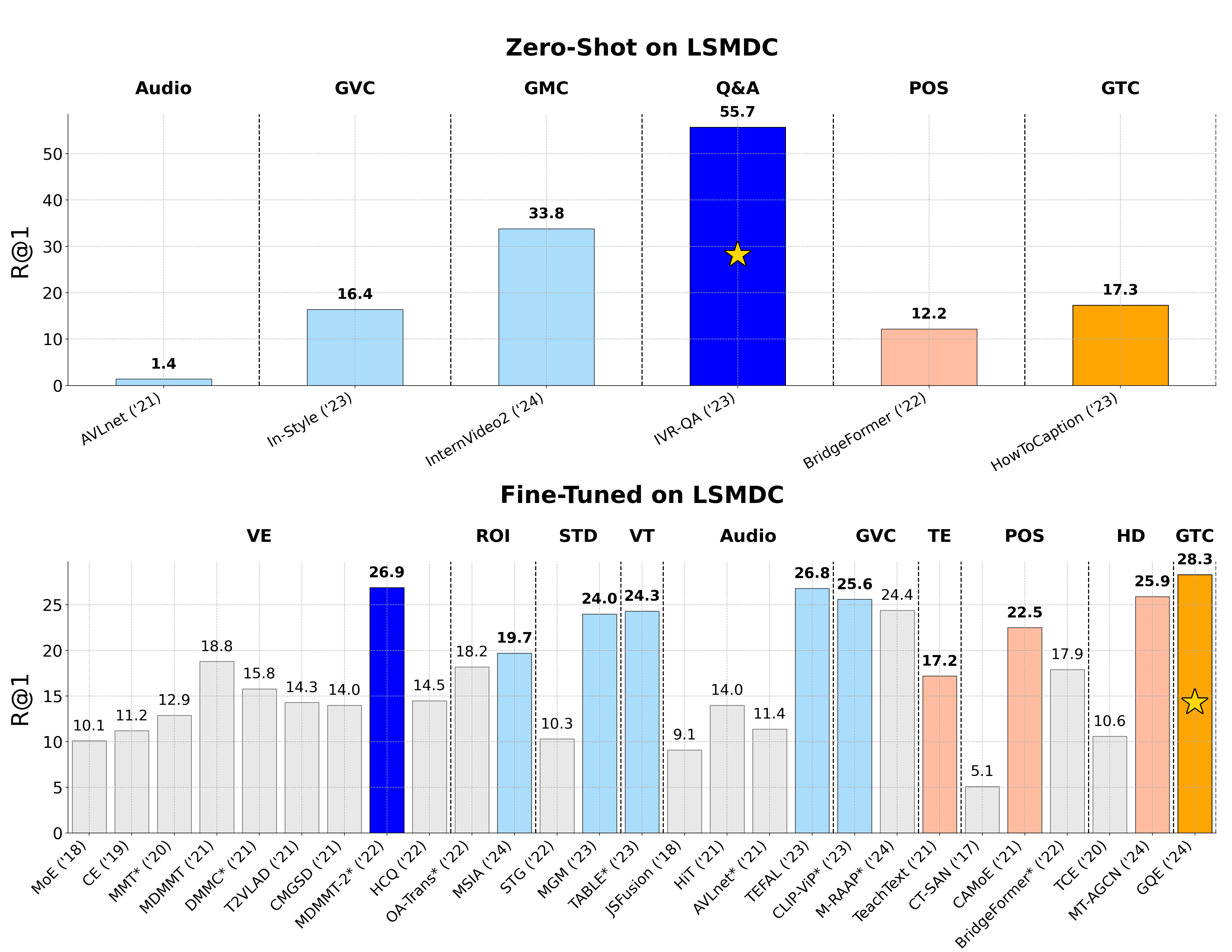}
    \end{minipage}
    \hfill
    \begin{minipage}[t]{0.48\linewidth}
        \centering
        \includegraphics[width=\linewidth]{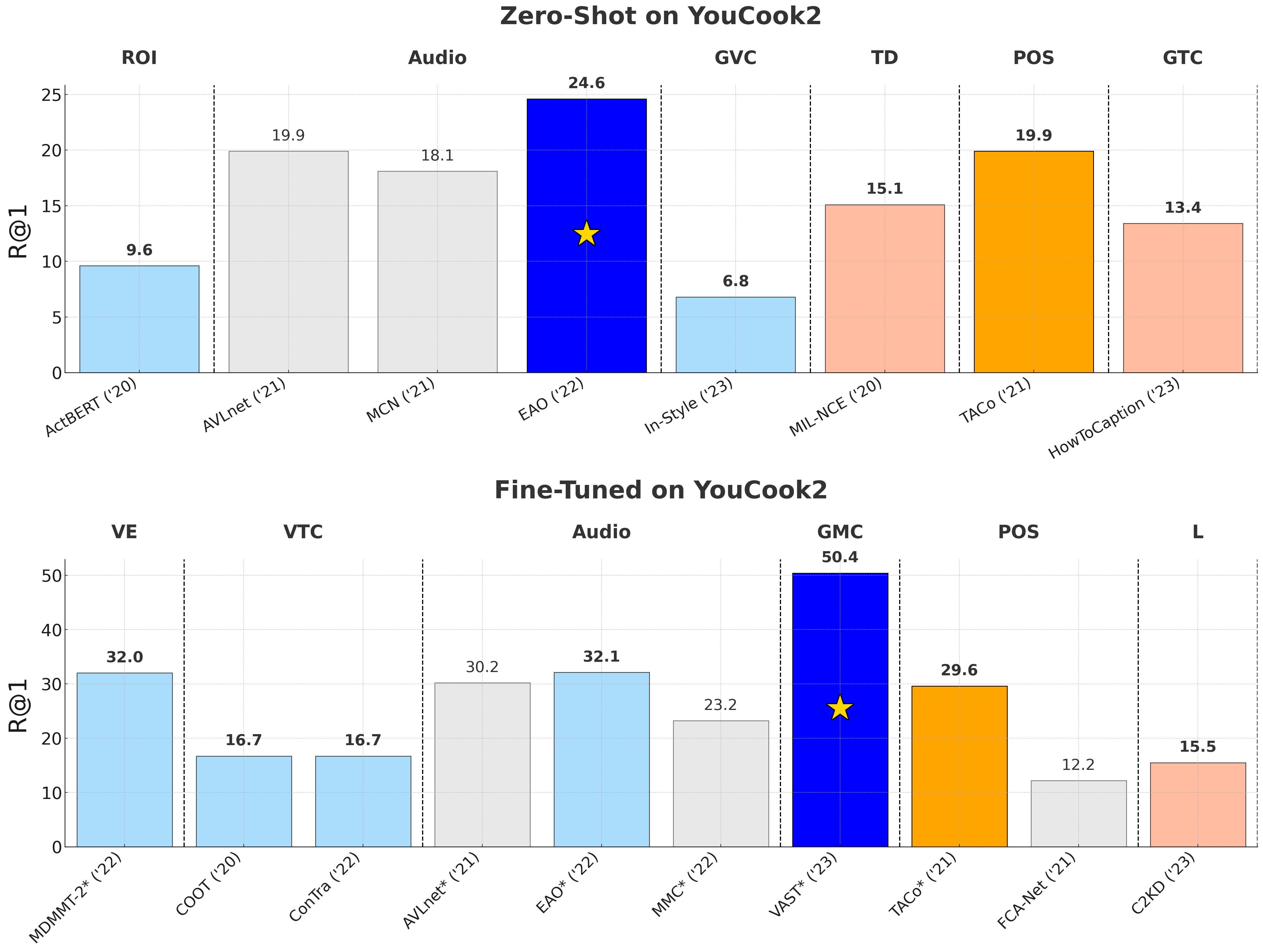}
    \end{minipage}
    	\caption{Performance comparison of Zero-Shot/Fine-Tuned models on LSMDC (left) and YouCook2 (right) across different types of auxiliary information. The \textcolor{gold}{\textbf{star}} absolute best-performing. \textcolor{blue}{\textbf{Blue}} and \textcolor{orange}{\textbf{orange}} bars best-performing across video- and text-extracted classes, respectively. \textcolor{lightskyblue}{\textbf{Light blue}} and \textcolor{lightsalmon}{\textbf{light salmon}} bars best-performing within video- and text-extracted classes, respectively. \textcolor{lightgray}{\textbf{Light grey}} other methods. $^*$: Pre-training. The class labels are defined in Tab.~\ref{tab:explanation_auxiliary_video} (video-extracted) and Tab.~\ref{tab:explanation_auxiliary_text} (text-extracted).}
        \label{fig:lsmdc_yc2_comparison}
\end{figure}

\paragraph{Results on LSMDC}
The conclusions highlighted in Fig.~\ref{fig:lsmdc_yc2_comparison} (left) further confirm similar trends observed in previous datasets, such as the dominance of Q$\&$A-based methods, i.e. IVR-QA (Q$\&$A) with $R@1=55.7$, outperforming InternVideo2 (GMC) and In-Style (GVC) by a large margin in the Zero-Shot setting. On the contrary, in the Fine-Tuned setting, the text-information method GQE~\cite{bai2025bridging} (GTC) achieves the highest $R@1$ of $28.3$ across all methods. Followed by MDMMT-2 (VE) and TEFAL~\cite{DBLP:conf/iccv/IbrahimiSWGSO23} (Audio) that demonstrate the importance of Audio in this domain-specific dataset (i.e movie domain).

\paragraph{Results on YouCook2.} 
Fig.~\ref{fig:lsmdc_yc2_comparison} (right) showcases the importance of Audio also for YouCook2. EAO~\cite{DBLP:conf/cvpr/ShvetsovaCR0KFH22} (Audio) outperforms all the Zero-Shot methods by achieving the best $R@1$ at $24.6$. However, VAST (GMC), with an $R@1$ of $50.4$, significantly outperforms all the other Fine-Tuned approaches, including the Audio-based method EAO (Audio) and the best text-information method, TACo~\cite{DBLP:conf/iccv/YangBG21} (POS), by a large margin of $+18.3$ and $+20.8$, respectively. These results demonstrate the importance of Audio in improving video-text alignment: on one hand through VAST (GMC), which incorporates Audio both in the caption generation process and as input modality of the model, and on the other hand through EAO (Audio), which combines Audio with all the possible feature combinations involving video and text.

\begin{figure}[t]
    \centering
    \begin{minipage}[t]{0.48\linewidth}
        \centering
        \includegraphics[width=\linewidth]{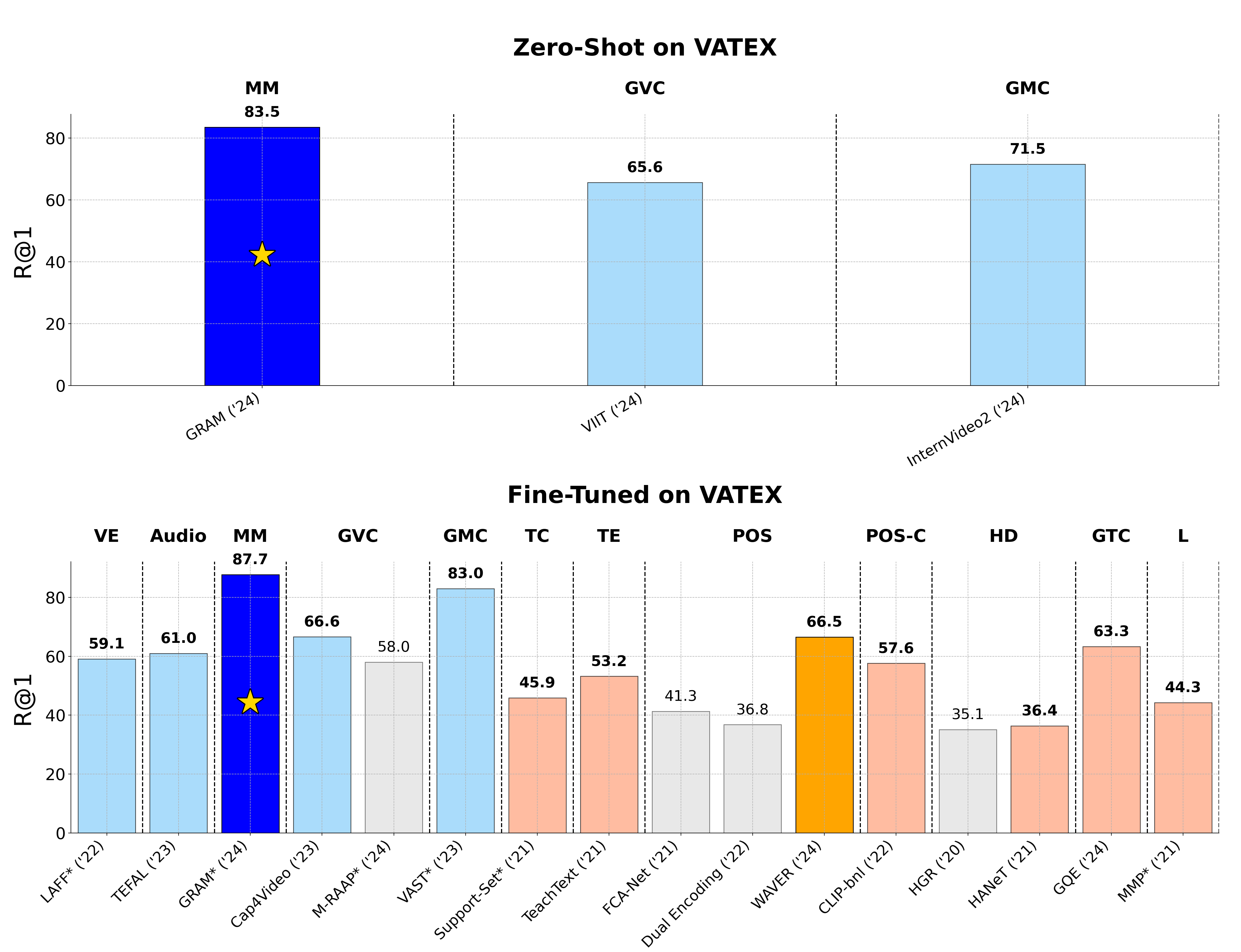}
    \end{minipage}
    \hfill
    \begin{minipage}[t]{0.48\linewidth}
        \centering
        \includegraphics[width=\linewidth]{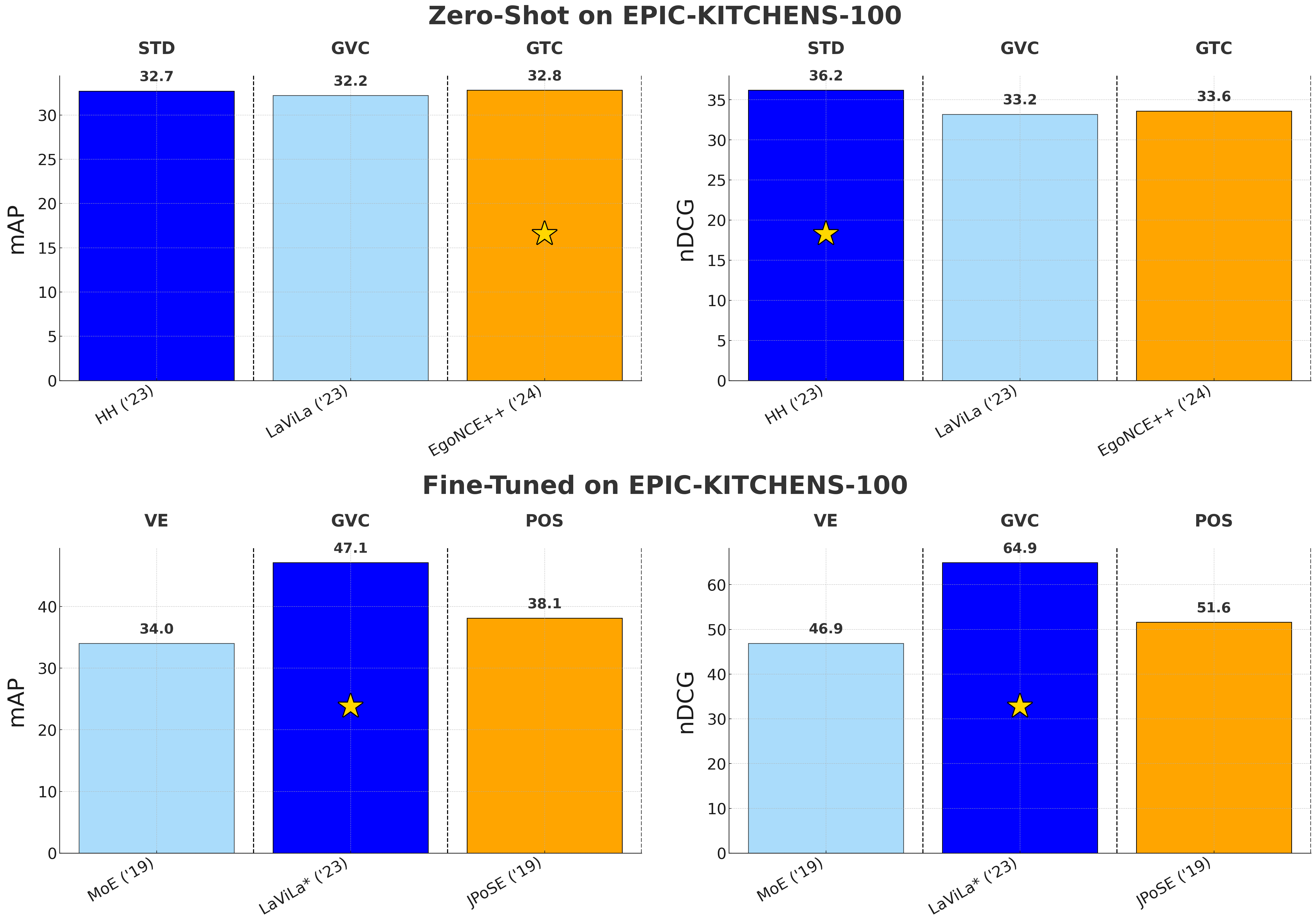}
    \end{minipage}
    	\caption{Performance comparison of Zero-Shot/Fine-Tuned models on VATEX (left) and EPIC-KITCHENS-100 (right) across different types of auxiliary information. The \textcolor{gold}{\textbf{star}} absolute best-performing. \textcolor{blue}{\textbf{Blue}} and \textcolor{orange}{\textbf{orange}} bars best-performing across video- and text-extracted classes, respectively. \textcolor{lightskyblue}{\textbf{Light blue}} and \textcolor{lightsalmon}{\textbf{light salmon}} bars best-performing within video- and text-extracted classes, respectively. \textcolor{lightgray}{\textbf{Light grey}} other methods. $^*$: Pre-training. The class labels are defined in Tab.~\ref{tab:explanation_auxiliary_video} (video-extracted) and Tab.~\ref{tab:explanation_auxiliary_text} (text-extracted).}
        \label{fig:epic_vatex_comparison}
\end{figure}

\paragraph{Results on VATEX} 
In VATEX, as shown in Fig.~\ref{fig:epic_vatex_comparison} (left), GRAM (MM) dominates the Zero-Shot setting with $R@1=83.5$, outperforming InternVideo2 (GMC) and VIIT (GVC). In the Fine-Tuned setting, GRAM (MM) again surpasses, with $R@1=87.7$, strong video-information methods such as VAST (GMC) and Cap4Video (GVC), which achieve best performance in their auxiliary class with $R@1=83.0$ and $R@1=66.6$, respectively. Although the top text-information methods, such as WAVER (POS) and GQE (GTC), fall significantly behind VAST (GMC), they show promising performance with $R@1=66.5$ and $R@1=63.3$, respectively. 

\paragraph{Results on EPIC-KITCHENS-100.} 
Fig.~\ref{fig:epic_vatex_comparison} (right) presents the multi-instance text-to-video results on the egocentric EPIC-KITCHENS-100 dataset. In the Zero-Shot scenario, HH~\cite{DBLP:conf/iccv/Zhang0Z23a} (STD) achieves the highest $nDCG=36.2$ and the second highest $mAP=32.7$, and EgoNCE++~\cite{DBLP:conf/iclr/XuWDSZJ25} (GTC) achieves the highest $mAP=32.8$ and the second top performance with $nDCG=36.2$. These results indicate that HH (STD) better ranks relevant clips highly (reflected in a higher $nDCG$, which is ranking-sensitive) through fine-grained spatial-temporal modelling. In contrast, EgoNCE++ (GTC) improves overall retrieval accuracy (reflected in higher $mAP$) by capturing global semantic relationships.
In the Fine-Tuned setting, LaViLa~\cite{DBLP:conf/cvpr/0006MKG23} (GVC) dominates all the other methods with $nDCG=64.9$ and $mAP=47.1$.

\paragraph{Conclusion and Findings.}

Overall, this analysis reveals regular patterns across datasets. In the Zero-Shot setting, Q$\&$A-based approaches achieve the best performance across MSR-VTT, MSVD, ActivityNet Captions, DiDeMo, and LSMDC. Meanwhile, video-information methods, particularly those leveraging generated video captions (GVC) or multi-modal signals (MM), consistently outperform text-information methods across most datasets. 

In the Fine-Tuned setting, while strong video-based approaches often dominate (e.g. VAST on ActivityNet Captions, DiDeMo, and YouCook2), several text-information methods, particularly MQVR (TC) and GQE (GTC), show competitive or even leading performance on datasets such as MSR-VTT, MSVD, and LSMDC. Furthermore, the importance of auxiliary modalities such as Audio becomes evident in domain-specific datasets such as YouCook2 and LSMDC. 

Altogether, these findings underline the crucial role of generated and multi-modal information in improving video retrieval performance, setting new higher thresholds for improvement across multiple benchmark datasets. 

\subsubsection{Overall Analysis.} 

We now extend the comparison between multiple datasets to identify interesting insights across video- and text-information methods. This comparison highlights some interesting findings that may lay the foundation for promising future directions in this field. 

\paragraph{Multiple Available Captions.} 

In trimmed datasets annotated with multiple captions per clip (e.g. MSR-VTT, MSVD, LSMDC and VATEX), text-information methods achieve competitive and leading performance. The availability of multiple human-labelled captions per clip provides a unique opportunity to extract more detailed and scaled information.
In particular, MQVR (TC) achieves the highest results on MSR-VTT and MSVD in the Fine-Tuned setting by aggregating multiple captions during training and testing. 

Many text-information methods leverage these multiple captions to extract more textual information, such as more POS tags, per clip. More precisely, in the Fine-Tuned setting, WAVER (POS) achieves the best performance on VATEX among all the text-information methods by extracting nouns and verbs to construct a Video Content Dictionary used in training. The more captions per clip, the more informative this dictionary becomes. 

Similar conclusions are observed for BridgeFormer~\cite{DBLP:conf/cvpr/GeGLLSQL22} (POS), which achieves top performance in its auxiliary class on MSVD by leveraging POS features for its multiple-choice question task, and MT-AGCN~\cite{DBLP:journals/mms/LvSN24} (HD) on MSVD and LSMDC, by leveraging semantic dependencies between actions and entities extracted via POS. Overall, these results highlight that increasing the amount of textual descriptions per clip significantly increases the effectiveness of these text-information methods.

\paragraph{Audio in Domain-Specific Datasets.}
In domain-specific datasets such as LSMDC (movie domain) and YouCook2 (cooking domain), Audio appears as a valuable additional modality, easy to extract, and fundamental for distinguishing fine-grained visual content. Audio-based architectures perform strongly on these three datasets, where speech, ambient sounds and cooking noises provide discriminative details that complement video and text. Relevant works that leverage Audio exclusively as input modality, without employing generative models or other modalities, include TEFAL (Audio), which reaches top results on LSMDC across all the Audio-methods in the Fine-Tuned setting, and EAO (Audio), which obtains the highest performance on YouCook2 in the Zero-Shot setting scenario and competitive results in the Fine-Tuned setting.

\paragraph{Temporal Context in Untrimmed Datasets.}
Capturing the temporal context of clips or objects in an untrimmed video is a promising resource, as demonstrated by COOT (VTC) on ActivityNet Captions and HH (STD) on EPIC-KITCHENS-100. Both methods model temporal dependencies between clips/sentences or objects/hands, respectively. Since ActivityNet Captions and EPIC-KITCHENS-100 consist of untrimmed videos, the temporal information naturally provided in these videos is essential to enrich video and/or text representations.

\paragraph{Alignment and Generation with Multiple Modalities.} 
Additional modalities have been used in recent works by proposing new modality alignment strategies, such as GRAM (MM), or by leveraging generative models, such as InternVideo2 (GMC) and VAST (GMC).

More precisely, GRAM (MM) introduces a novel geometric alignment loss to simultaneously align multiple modalities, achieving leading performance on MSR-VTT, DiDeMo, ActivityNet Captions, and VATEX. In parallel, InternVideo2 (GMC) and VAST (GMC) enhance multi-modal alignment by fusing generated visual, audio, and speech captions using LLMs, and embedding these modalities into their architectures through dedicated alignment mechanisms. These approaches demonstrate top performance across MSR-VTT, DiDeMo, ActivityNet Captions, VATEX, and LSMDC, highlighting the effectiveness of multi-modal and generative strategies for improving multi-modal alignment.

\paragraph{Information Generated by Generative Models.} 

Recent research has highlighted the impact of \emph{generative models}, such as video captioners, VLMs and LLMs. These approaches have established new benchmarks across multiple datasets by leveraging video- and text-extracted information, including additional captions generated from videos (GVC), refined/paraphrased captions (GTC), and chat-based systems (Q$\&$A).

Q$\&$A-based methods generate pairs of questions and answers to enrich the contextual details available for videos. Approaches such as IVR-QA (Q$\&$A) and MERLIN (Q$\&$A) have demonstrated strong performance in the Zero-Shot setting, achieving state-of-the-art results on different dataset domains such as MSR-VTT, MSVD, ActivityNet Captions, DiDeMo, and LSMDC.

Another line of research focuses on generating auxiliary captions from video (GVC) and refining text (GTC). GVC-based methods achieve strong performances across many datasets, such as MSR-VTT, MSVD, DiDeMo, ActivityNet Captions, LSMDC, VATEX, and EPIC-KITCHENS-100, by generating high-quality captions from the video content using video captioners or VLMs. Among these methods, T2VIndexer (GVC) shows promising results on MSR-VTT and MSVD, CLIP-ViP~\cite{DBLP:conf/iclr/XueS0FSLL23} (GVC) obtains high performance on LSMDC, Cap4Video (GVC) on VATEX and LaViLa (GVC) on EPIC-KITCHENS-100 in the Fine-Tuned setting.

In parallel, GTC-based methods such as HowToCaption (GTC), GQE (GTC), and EgoNCE++ (GTC) use the original captions to generate new refined textual descriptions via LLMs. Compared to all the text-information methods, HowToCaption (GTC) performed at the top in MSR-VTT, MSVD, and LSMDC in the Zero-Shot. However, its performance is still far from that of video-information approaches. GQE (GTC) achieves strong fine-tuned performance on MSR-VTT and VATEX and state-of-the-art results on LSMDC. While EgoNCE++ performs well in the zero-shot setting on EPIC-KITCHENS-100.

Both GVC- and GTC-based approaches show consistent improvements across datasets. However, GVC-based methods generally outperform GTC-based methods in terms of the number of datasets. GVC-based methods outperform GTC-based methods across $3$ datasets: MSR-VTT and MSVD in both settings, and VATEX in the Fine-Tuned setting. On the contrary, GTC-based methods perform better on YouCook2 in the Zero-Setting Scenario and LSMDC in both scenarios.

\paragraph{Conclusion and Findings.}

In conclusion, the overall analysis highlights several consistent trends. Datasets annotated with multiple captions, such as MSR-VTT, MSVD, and VATEX, strongly benefit text-information methods. Audio-based approaches remain particularly effective on domain-specific datasets like LSMDC and YouCook2, and temporal context is fundamental for untrimmed video datasets such as ActivityNet Captions and EPIC-KITCHENS-100. Furthermore, multi-modal alignment strategies (MM) and the use of generative models (GVC, GMC, GTC, Q$\&$A) consistently achieve strong performance, with GVC- and GMC-based methods generally outperforming text-refinement approaches (GTC). These new directions are reshaping this area and establishing strong new baselines, pointing towards future research directions centred on richer auxiliary information integration.

\section{Future Research}\label{sec6}
Despite the significant advancements in Text-to-Video retrieval methods through the utilisation of auxiliary information extracted from both videos and texts, as shown in Section~\ref{sec5}, there are still several promising research directions that might be explored. These directions not only aim to enhance retrieval performance but also redefining and expanding the retrieval task itself.

\begin{enumerate}
    \item \textbf{Modernisation of Text-to-Video Retrieval}: Most of the papers studied in this review focus on improving the retrieval performance by introducing different auxiliary information. However, some attempt to redefine the retrieval task itself. Approaches such as multi-query strategies~\cite{DBLP:conf/eccv/WangWNR22} or Q$\&$A systems~\cite{DBLP:conf/iccv/LiangA23,DBLP:conf/emnlp/HanPLLK24} have demonstrated superior performance across various benchmark datasets, highlighting the importance of moving beyond traditional retrieval settings.
    \item \textbf{Innovative Modality Alignment Techniques}: Current efforts, such as GRAM and LanguageBind, represent pioneering attempts to redefine multimodal learning by introducing geometric multimodal spaces or using language as an anchor modality to align all the others. Similarly, VAST and InternVideo2 align different modalities not only within their architecture design but also through an innovative caption generation pipeline. 
    \item \textbf{Limited Metadata in Existing Datasets}: Few datasets (e.g., Ego4D~\cite{DBLP:conf/cvpr/GraumanWBCFGH0L22}) include additional metadata beyond audio. Future benchmark datasets incorporating information from metadata (e.g. hashtags) available on public video platforms such as YouTube can provide additional contextual information and offer a better understanding of video content. 
    \item \textbf{Shortage of Multi-lingual Datasets}: English-centric datasets limit multilingual retrieval introducing a significant language bias. Although datasets like Epic-Kitchens-100~\cite{Damen2018EPICKITCHENS} and VATEX~\cite{DBLP:conf/iccv/WangWCLWW19} include manual annotated multilingual captions, their potential remains under-explored. Addressing this limitation is crucial because real-world scenarios often involve multilingual queries. Each language has a unique vocabulary, grammar, and morphology that can provide valuable context, enhancing the alignment between text and video as shown from a few works~\cite{DBLP:conf/naacl/HuangPHNMH21,DBLP:conf/icassp/RouditchenkoCSTFKKHKG23,DBLP:journals/ir/MadasuASRTBL23} by using state-of-the-art machine translation methods. 
    \item \textbf{Using Visual-Language Models (VLMs) and Large Language Models (LLMs) for Video Retrieval}: Recent progress in (VLMs)~\cite{li2023blip,liu2024visual,zhang2023video,cheng2024videollama,damonlpsg2023videollama} and (LLMs)~\cite{touvron2023llama,touvron2023llama2,vicuna2023,dubey2024llama} has enhanced Text-to-Video retrieval performance as we have shown in Sec.~\ref{sec5}. These models excel in understanding, combining, and extracting auxiliary information from videos and captions, leading to high retrieval performance on standard benchmarks as shown in~\cite{DBLP:conf/mm/Li0GL0024, DBLP:conf/cvpr/WuLFWO23, bai2025bridging, DBLP:conf/eccv/ShvetsovaKHRSK24,DBLP:conf/cvpr/0006MKG23}. 
    
    \item \textbf{More Captions, More Perspectives}: Generative models can create alternative video descriptions, improving retrieval through varied textual perspectives.
    Few works~\cite{DBLP:conf/mm/Li0GL0024, DBLP:conf/cvpr/WuLFWO23, bai2025bridging, DBLP:conf/eccv/ShvetsovaKHRSK24,DBLP:conf/cvpr/0006MKG23,DBLP:conf/nips/ChenLWZSZL23,DBLP:conf/eccv/WangLLYHCPZWSJLXZHQWW24} lay the foundations of generating refined or auxiliary descriptions that offer different textual perspectives—whether in terms of detail or stylistic variation. Promising results have been shown by few works~\cite{DBLP:conf/nips/ChenLWZSZL23,DBLP:conf/eccv/WangLLYHCPZWSJLXZHQWW24} that propose to extract modality-specific captions that are then merged in a single final caption via LLM. Future research could investigate further how incorporating multiple, stylistically diverse descriptions can lead to better alignment between modalities and, consequently, improved retrieval performance.

\end{enumerate}
In conclusion, the future of Text-to-Video retrieval with auxiliary information is rich in opportunities. By addressing these key research directions—ranging from paradigm shifts and innovative modality alignment to multilingual inclusivity and advanced modalities and model integration—the field can achieve significant improvements in both performance and applicability.

\section{Conclusion}\label{sec7}
This review has explored Text-to-Video retrieval methodologies that leverage auxiliary information to address the semantic gap between video and text modalities. By categorising these approaches based on the source (video or text) and type (visual, other modalities, or generated) of auxiliary information, we highlighted their role in enhancing alignment and improving retrieval performance. Additionally, we reviewed their results on widely used benchmark datasets, highlighting the achievements of current methods.
The integration of auxiliary information has proven to be a powerful tool in this field. Different techniques and auxiliary information have been studied in this review from methods that leverage pre-trained models for extracting visual and audio cues to generative approaches for creating captions and metadata. All these methods have shown an effort to create more robust and context-aware retrieval methods. However, several challenges remain, presenting opportunities for further research. By identifying these trends, challenges, and directions, this survey provides an overview for advancing the field of Text-to-Video retrieval, aiming to inspire innovative solutions that leverage auxiliary information to overcome modality gaps and deliver more effective and inclusive retrieval systems.



\bibliographystyle{ACM-Reference-Format}
\bibliography{sample-base}
\newpage

\appendix

\section{Datasets.}
\label{appendix_datasets}


\subsection{Pre-Training Datasets}
These datasets are used for pre-training because they are the largest ones available for both third-person (exocentric) and first-person (egocentric) perspectives.
\noindent \textbf{HowTo100M}~\cite{DBLP:conf/iccv/MiechZATLS19} is a large-scale dataset widely used for pre-training. It comprises $136$M video clips paired with automatically transcribed captions from approximately $1.2$M YouTube instructional videos. Due to automatic extraction, captions are not always well aligned with the video content. \textit{Multi-HowTo100M}~\cite{DBLP:conf/naacl/HuangPHNMH21} is a variant with $1.09$M videos featuring multi-lingual subtitles in $8$ languages and \textit{HowToCaption}~\cite{DBLP:conf/eccv/ShvetsovaKHRSK24} is another variant consisting of $1.2$M instructional videos with ASR subtitles refined into proper captions using the HowToCaption method and Vicuna-13B~\cite{vicuna2023}.
\textbf{Ego4D}~\cite{DBLP:conf/cvpr/GraumanWBCFGH0L22} is a large-scale egocentric dataset. It comprises $3,670$ hours of videos from $923$ individuals across $74$ global locations, showcasing daily activities and interactions. \textit{EgoClip}~\cite{DBLP:conf/nips/LinWSWYXGTZKCWD22}, is an extension that offers $3.85$M clip-caption pairs with dense, localised temporal annotations.

\subsection{Downstream Datasets}
These datasets are much smaller in scale and are utilised as downstream benchmarks for Text-to-Video retrieval.
\textbf{MSR-VTT}~\cite{DBLP:conf/cvpr/XuMYR16} is a benchmark dataset for video-language tasks. It includes $10$k YouTube video clips. Each video has $20$ unique captions and lasts $10$–$30$ seconds. A multi-lingual version, \textit{Multi-MSRVTT}, translates the original English captions into $8$ languages~\cite{DBLP:conf/naacl/HuangPHNMH21}.
We consider the split \textit{1k-A} introduced in~\cite{DBLP:conf/eccv/YuKK18}.
\textbf{MSVD}~\cite{DBLP:conf/acl/ChenD11} is a popular benchmark for video-language tasks. It comprises $1,970$ short YouTube video clips of everyday activities. 
Originally annotated in different languages, English has always been used in retrieval. However, it has been officially expanded to other languages~\cite{DBLP:journals/mt/CitamakCKEEMS21,DBLP:journals/corr/abs-2306-11341}.
\textbf{ActivityNet Captions}~\cite{DBLP:conf/iccv/KrishnaHRFN17} is a benchmark dataset for dense video captioning and retrieval, featuring over $20$k annotated video segments from approximately $15$k YouTube videos. It provides natural language captions with precise temporal boundaries. This dataset is commonly used in the P2V retrieval task.
\textbf{DiDeMo}~\cite{DBLP:conf/iccv/HendricksWSSDR17} is a benchmark dataset for video retrieval and temporal video grounding. It contains approximately $10$k Flickr videos and over $40$k human-annotated descriptions. Each video is segmented into $5$-second intervals with precise temporal annotations. This dataset is commonly used in the P2V setting.
\textbf{LSMDC}~\cite{DBLP:conf/cvpr/RohrbachRTS15} comprises $118$k text-video pairs from $202$ movies, totalling $158$ hours of video. Clips  include rich natural language descriptions of actions, events, and character interactions.
\textbf{YouCook2}~\cite{DBLP:conf/aaai/ZhouXC18} is a retrieval dataset in the cooking domain. It contains $2$k untrimmed YouTube videos. A multi-language version, \textbf{Multi-YouCook2}, was introduced in~\cite{DBLP:conf/icassp/RouditchenkoCSTFKKHKG23}, with captions translated into $8$ languages.
\textbf{VATEX}~\cite{DBLP:conf/iccv/WangWCLWW19} is a multi-lingual (Chinese and English) video-text dataset containing $35,000$ videos. Each $10$-second video has five paired and five unpaired descriptions.
\textbf{Epic-Kitchens-100}~\cite{Damen2018EPICKITCHENS} is a $100$-hour egocentric video dataset from $37$ participants in kitchen settings, featuring over $90,000$ action segments with fine-grained verb-noun annotations for detailed action-object analysis. Captions are reported in four different languages: English, Italian, Greek, or Chinese. Non-English captions are translated to English.

\section{Full Tables of Results}
\label{full_table}
\begin{table*}[ht!]
\centering
\resizebox{0.6\linewidth}{!}{%
\begin{tabular}{llcccccc}
&\multicolumn{1}{l}{Method}
&\multicolumn{1}{c}{Year}
&\multicolumn{1}{c|}{Auxiliary Info$^{\dagger}$}
&\multicolumn{1}{c}{R@1$\uparrow$}
&\multicolumn{1}{c}{R@5$\uparrow$}
&\multicolumn{1}{c}{R@10$\uparrow$}
&\multicolumn{1}{c}{MR$\downarrow$}\\\hline

\multirow{51}{*}{\centering Fine-Tuning (ft)}
&CE~\cite{DBLP:conf/bmvc/LiuANZ19}&2019&VE&20.9&48.8&62.4&6\\
&MMT$^{*}$~\cite{DBLP:conf/eccv/Gabeur0AS20}&2020&VE&26.6&57.1&69.6&4\\
&T2VLAD~\cite{DBLP:conf/cvpr/WangZ021}&2021&VE&29.5&59.0&70.1&4\\
&DMMC$^{*}$~\cite{DBLP:conf/ijcai/WangZCCZPGWS21}&2021&VE&31.2&62.8&76.4&3\\
&CMGSD~\cite{DBLP:conf/sigir/HeWFJLZT21}&2021&VE&26.1&56.8&69.7&4\\
&MDMMT~\cite{DBLP:conf/cvpr/DzabraevKKP21}&2021&VE&38.9&69.0&79.7&2\\
&LAFF$^{*}$~\cite{DBLP:conf/eccv/HuCWZDL22}&2022&VE&42.6&71.8&81.0&-\\
&TxV~\cite{DBLP:conf/eccv/GalanopoulosM22}&2022&VE&36.5&66.9&77.7&2\\
&HCQ~\cite{DBLP:conf/www/WangCLZLXX22}&2022&VE&25.9&54.8&69.0&5\\
&MDMMT-2$^{*}$~\cite{DBLP:journals/corr/abs-2203-07086}&2022&VE&48.5&75.4&83.9&2\\\cline{2-8}

&JSFusion~\cite{DBLP:conf/eccv/YuKK18}&2018&Audio&10.2&31.2&43.2&13\\
&AVLnet$^{*}$~\cite{DBLP:conf/interspeech/RouditchenkoBHC21}&2021&Audio&22.5&50.5&64.1&5\\
&HiT$^{*}$~\cite{DBLP:conf/iccv/Liu0QCDW21}&2021 &Audio&30.7&60.9&73.2&3\\
&EAO$^{*}$~\cite{DBLP:conf/cvpr/ShvetsovaCR0KFH22}&2022&Audio&23.7&52.1&63.7&4\\
&MMC$^{*}$~\cite{DBLP:conf/wacv/GabeurN0AS22}&2022&Audio&28.7&59.5&70.3&4\\
&LAVMIC$^{*}$~\cite{DBLP:conf/eccv/NagraniSSHMSS22}&2022&Audio&35.8&65.1&76.9&-\\
&TEFAL~\cite{DBLP:conf/iccv/IbrahimiSWGSO23}&2023&Audio&49.4&75.9&83.9&2\\\cline{2-8}

&TACo$^{*}$~\cite{DBLP:conf/iccv/YangBG21}&2021&POS&26.7&54.5&68.2&4\\
&CAMoE~\cite{DBLP:journals/corr/abs-2109-04290}&2021&POS&44.6&72.6&81.8&2\\
&FCA-Net~\cite{DBLP:conf/mm/HanCXZZC21}&2021&POS&23.2&55.6&70.3&3\\
&BridgeFormer$^{*}$~\cite{DBLP:conf/cvpr/GeGLLSQL22}&2022&POS&37.6&64.8&75.1&3\\
&Dual Encoding~\cite{DBLP:journals/pami/DongLXYYWW22}&2022&POS&21.1&48.7&60.2&6\\
&WAVER~\cite{DBLP:conf/icassp/LeK0L24}&2024&POS&47.8&74.6&83.9&2\\\cline{2-8}

&ActBERT$^{*}$~\cite{DBLP:conf/cvpr/ZhuY20a}&2020&ROI&16.3&42.8&56.9&10\\
&OA-Trans$^{*}$~\cite{DBLP:conf/cvpr/WangGCY0SQS22}&2022&ROI&35.8&63.4&76.5&3\\
&MSIA~\cite{chen2024multilevel}&2024&ROI&41.6&68.8&79.8&2\\\cline{2-8}

&STG~\cite{DBLP:journals/tmm/SongCWJ22}&2022&STD&15.5&39.2&50.4&10\\
&MGM~\cite{DBLP:conf/mm/SongCJ23}&2023&STD&47.2&74.6&83.0&2\\
&BiC-Net~\cite{DBLP:journals/tomccap/HanZSXCC24}&2024&STD&39.4&75.5&86.7&2\\\cline{2-8}

&TCE~\cite{DBLP:conf/sigir/YangD0W0C20}&2020&HD&16.1&38.0&51.5&10\\
&SHE-Net~\cite{DBLP:journals/corr/abs-2404-14066}&2024&HD&45.3&74.9&84.2&2\\
&MT-AGCN~\cite{DBLP:journals/mms/LvSN24}&2024&HD&48.1&73.6&83.1&2\\\cline{2-8}

&Support-Set$^{*}$~\cite{DBLP:conf/iclr/Patrick0AMHHV21}&2021&TC&30.1&58.5&69.3&3\\
&MQVR~\cite{DBLP:conf/eccv/WangWNR22}&2022&TC&73.1&94.1&97.8&1\\
&LINAS~\cite{DBLP:conf/eccv/FangWZHH22}&2022&TC&27.1&59.8&71.7&4\\\cline{2-8}

&SEA~\cite{DBLP:journals/tmm/0001ZXJ021}&2021&TE&23.8&50.3&63.8&5\\
&TeachText~\cite{DBLP:conf/iccv/CroitoruBLJZAL21}&2021&TE&29.6&61.6&74.2&3\\\cline{2-8}

&MMP$^{*}$~\cite{DBLP:conf/naacl/HuangPHNMH21}&2021&L&23.8&52.6&65.0&-\\
&MuMUR~\cite{DBLP:journals/ir/MadasuASRTBL23}&2023&L&46.6&72.6&82.2&2\\
&C2KD~\cite{DBLP:conf/icassp/RouditchenkoCSTFKKHKG23}&2023&L&26.4&-&-&-\\\cline{2-8}

&HiSE~\cite{DBLP:conf/mm/WangXHLJHD22}&2022&VT&45.0&72.7&81.3&2\\
&ALPRO$^{*}$~\cite{DBLP:conf/cvpr/Li0LNH22}&2022&VT&33.9&60.7&73.2&3\\
&TABLE$^{*}$~\cite{DBLP:conf/aaai/ChenWLQMS23}&2023&VT&47.1&74.3&82.9&2\\\cline{2-8}

&CLIP-bnl~\cite{DBLP:conf/mm/WangCH022}&2022&POS-C&42.1&68.4&79.6&-\\\cline{2-8}

&VAST$^{*}$~\cite{DBLP:conf/nips/ChenLWZSZL23}&2023&GMC&63.9&84.3&89.6&-\\\cline{2-8}

&GRAM$^{*}$~\cite{DBLP:conf/iclr/CicchettiGSC25}&2024&MM&64.0&-&89.3&-\\\cline{2-8}

&CLIP-ViP$^{*}$~\cite{DBLP:conf/iclr/XueS0FSLL23}&2023&GVC&50.1&74.8&84.6&1\\
&Cap4Video~\cite{DBLP:conf/cvpr/WuLFWO23}&2023&GVC&49.3&74.3&83.8&2\\
&M-RAAP$^{*}$~\cite{DBLP:conf/sigir/DongFZYYG24}&2024&GVC&46.1&70.4&80.6&-\\
&T2VIndexer$^{*}$~\cite{DBLP:conf/mm/Li0GL0024}&2024&GVC&55.1&77.2&85.0&-\\\cline{2-8}

&GQE~\cite{bai2025bridging}&2024&GTC&52.1&76.8&86.3&1\\
\hline\hline
\end{tabular}}
\caption{Sentence-to-Clip results on MSR-VTT (Fine-Tuning). $^{*}$: Pre-training. $^{\dagger}$: see Tabs~\ref{tab:explanation_auxiliary_video} -~\ref{tab:explanation_auxiliary_text} for auxiliary info.}
\label{tab:msrvtt_results_ft}
\end{table*}

\begin{table*}[ht!]
\centering
\resizebox{0.6\linewidth}{!}{%
\begin{tabular}{llcccccc}
&\multicolumn{1}{l}{Method}
&\multicolumn{1}{c}{Year}
&\multicolumn{1}{c|}{Auxiliary Info$^{\dagger}$}
&\multicolumn{1}{c}{R@1$\uparrow$}
&\multicolumn{1}{c}{R@5$\uparrow$}
&\multicolumn{1}{c}{R@10$\uparrow$}
&\multicolumn{1}{c}{MR$\downarrow$}\\\hline

\multirow{19}{*}{\centering Zero-Shot (zs)}
&ActBERT~\cite{DBLP:conf/cvpr/ZhuY20a}&2020&ROI&8.6&23.4&33.1&36\\
&OA-Trans~\cite{DBLP:conf/cvpr/WangGCY0SQS22}&2022&ROI&23.4&47.5&55.6&8\\\cline{2-8}

&AVLnet~\cite{DBLP:conf/interspeech/RouditchenkoBHC21}&2021&Audio&8.3&19.2&27.4&47\\
&VATT~\cite{DBLP:conf/nips/AkbariYQCCCG21}&2021&Audio&-&-&29.7&49\\
&MCN ~\cite{DBLP:conf/iccv/ChenRDK0BPKFHGP21}&2021&Audio&10.5&25.2&33.8&-\\
&LAVMIC~\cite{DBLP:conf/eccv/NagraniSSHMSS22}&2022&Audio&19.4&39.5&50.3&-\\
&EAO~\cite{DBLP:conf/cvpr/ShvetsovaCR0KFH22}&2022&Audio&9.3&22.9&31.2&35\\\cline{2-8}

&ALPRO~\cite{DBLP:conf/cvpr/Li0LNH22}&2022&VT&24.1&44.7&55.4&8\\\cline{2-8}

&BridgeFormer~\cite{DBLP:conf/cvpr/GeGLLSQL22}&2022&POS&26.0&46.4&56.4&7\\\cline{2-8}

&VAST~\cite{DBLP:conf/nips/ChenLWZSZL23}&2023&GMC&49.3&68.3&73.9&-\\
&InternVideo2~\cite{DBLP:conf/eccv/WangLLYHCPZWSJLXZHQWW24}&2024&GMC&55.9&78.3&85.1&-\\\cline{2-8}

&ImageBind~\cite{DBLP:conf/cvpr/GirdharELSAJM23}&2023&MM&36.8&61.8&70.0&-\\
&GRAM~\cite{DBLP:conf/iclr/CicchettiGSC25}&2024&MM&54.8&-&82.9&-\\
&LanguageBind~\cite{DBLP:conf/iclr/ZhuLNYCWPJZLZ0024}&2024&MM&44.8&70.0&78.7&-\\\cline{2-8}

&MIL-NCE~\cite{DBLP:conf/cvpr/MiechASLSZ20}&2020&TD&9.9&24.0&32.4&29\\\cline{2-8}

&HowToCaption~\cite{DBLP:conf/eccv/ShvetsovaKHRSK24}&2023&GTC&37.6&62.0&73.3&3\\\cline{2-8}

&IVR-QA~\cite{DBLP:conf/iccv/LiangA23}&2023&Q$\&$A&67.9&89.5&94.9&1\\\cline{2-8}

&In-Style~\cite{DBLP:conf/iccv/ShvetsovaKSK23}&2023&GVC&36.0&61.9&71.5&3\\
&VIIT~\cite{DBLP:conf/cvpr/00060ZWCMSA0GKY24}&2024&GVC&48.4&73.5&81.9&-\\\hline\hline
\end{tabular}}
\caption{Sentence-to-Clip results on MSR-VTT (Zero-Shot). $^{\dagger}$: see Tabs~\ref{tab:explanation_auxiliary_video} -~\ref{tab:explanation_auxiliary_text} for auxiliary info.}
\label{tab:msrvtt_results_zs}
\end{table*}

\begin{table*}[ht]
\centering
\resizebox{0.6\linewidth}{!}{%
\begin{tabular}{llcccccc}

&\multicolumn{1}{l}{Method}
&\multicolumn{1}{c}{Year}
&\multicolumn{1}{c}{Auxiliary Info$^{\dagger}$}
&\multicolumn{1}{c}{R@1$\uparrow$}
&\multicolumn{1}{c}{R@5 $\uparrow$}
&\multicolumn{1}{c}{R@10$\uparrow$}
&\multicolumn{1}{c}{MR$\downarrow$}\\\hline

\multirow{7}{*}{\centering Zero-Shot (zs)}
&BridgeFormer~\cite{DBLP:conf/cvpr/GeGLLSQL22}&2022&POS&43.6&74.9&84.9&2\\\cline{2-8}
&HowToCaption~\cite{DBLP:conf/eccv/ShvetsovaKHRSK24}&2023&GTC&44.5&73.3&82.1&2\\\cline{2-8}
&InternVideo2~\cite{DBLP:conf/eccv/WangLLYHCPZWSJLXZHQWW24}&2024&GMC&59.3&84.4&89.6&-\\\cline{2-8}
&In-Style~\cite{DBLP:conf/iccv/ShvetsovaKSK23}&2023&GVC&44.9&72.7&81.1&2\\\cline{2-8}
&LanguageBind~\cite{DBLP:conf/iclr/ZhuLNYCWPJZLZ0024}&2024&MM&53.9&80.4&87.8&-\\\cline{2-8}

&IVR-QA~\cite{DBLP:conf/iccv/LiangA23}&2023&Q$\&$A&74.2&93.4&97.9&1\\
&MERLIN~\cite{DBLP:conf/emnlp/HanPLLK24}&2024&Q$\&$A&77.6&94.5&97.3&-\\\hline\hline

\multirow{27}{*}{\centering Fine-Tuning (ft)}
&CE~\cite{DBLP:conf/bmvc/LiuANZ19}&2019&VE&19.8&49.0&63.8&6\\
&MFGATN~\cite{DBLP:conf/mir/HaoZWZLW21}&2021&VE&22.7&55.1&69.3&4\\
&ARFAN~\cite{DBLP:conf/icmcs/HaoZWZLWM21}&2021&VE&21.8&51.6&66.3&5\\
&LAFF$^{*}$~\cite{DBLP:conf/eccv/HuCWZDL22}&2022&VE&45.4&76.0&84.6&-\\
&MDMMT-2$^{*}$~\cite{DBLP:journals/corr/abs-2203-07086}&2022&VE&56.8&83.1&89.2&1\\\cline{2-8}

&ViSERN~\cite{DBLP:conf/ijcai/FengZG020}&2020&STD&18.1&48.4&61.3&6\\
&MGM~\cite{DBLP:conf/mm/SongCJ23}&2023&STD&45.7&76.4&85.6&2\\
&BiC-Net~\cite{DBLP:journals/tomccap/HanZSXCC24}&2024&STD&24.6&57.0&70.3&4\\\cline{2-8}

&Support-Set$^{*}$~\cite{DBLP:conf/iclr/Patrick0AMHHV21}&2021&TC&28.4&60.0&72.9&4\\
&CF-GNN~\cite{DBLP:journals/tmm/WangGYX21}&2021&TC&22.8&50.9&63.6&6\\
&MQVR~\cite{DBLP:conf/eccv/WangWNR22}&2022&TC&61.7&89.5&94.9&1\\
&LINAS~\cite{DBLP:conf/eccv/FangWZHH22}&2022&TC&-&57.6&-&4\\\cline{2-8}

&SEA~\cite{DBLP:journals/tmm/0001ZXJ021}&2021&TE&24.6&55.0&67.9&4\\
&TeachText~\cite{DBLP:conf/iccv/CroitoruBLJZAL21}&2021&TE&25.4&56.9&71.3&4\\\cline{2-8}

&CAMoE~\cite{DBLP:journals/corr/abs-2109-04290}&2021&POS&46.9&76.1&85.5&-\\
&BridgeFormer$^{*}$~\cite{DBLP:conf/cvpr/GeGLLSQL22}&2022&POS&52.0&82.8&90.0&1\\
&WAVER~\cite{DBLP:conf/icassp/LeK0L24}&2024&POS&50.2&83.5&88.1&2\\\cline{2-8}

&OA-Trans$^{*}$~\cite{DBLP:conf/cvpr/WangGCY0SQS22}&2022&ROI&39.1&68.4&80.3&2\\
&MSIA~\cite{chen2024multilevel}&2024&ROI&45.4&74.4&83.3&2\\\cline{2-8}

&HiSE~\cite{DBLP:conf/mm/WangXHLJHD22}&2022&VT&45.9&76.2&84.6&2\\
&TABLE$^{*}$~\cite{DBLP:conf/aaai/ChenWLQMS23}&2023&VT&49.9&79.3&87.4&2\\\cline{2-8}

&Cap4Video~\cite{DBLP:conf/cvpr/WuLFWO23}&2023&GVC&51.8&80.8&88.3&1\\
&T2VIndexer$^{*}$~\cite{DBLP:conf/mm/Li0GL0024}&2024&GVC&54.0&81.3&88.3&-\\\cline{2-8}

&MuMUR~\cite{DBLP:journals/ir/MadasuASRTBL23}&2023&L&47.2&77.3&86.2&2\\\cline{2-8}

&SHE-Net~\cite{DBLP:journals/corr/abs-2404-14066}&2024&HD&47.6&76.8&85.5&2\\
&MT-AGCN~\cite{DBLP:journals/mms/LvSN24}&2024&HD&50.9&79.1&87.7&2\\\cline{2-8}

&GQE~\cite{bai2025bridging}&2024&GTC&51.1&81.1&88.6&1\\\hline\hline
\end{tabular}}
\caption{Sentence-to-Clip results on MSVD. $^{*}$: Pre-training. $^{\dagger}$: see Tabs~\ref{tab:explanation_auxiliary_video} -~\ref{tab:explanation_auxiliary_text} for the type of auxiliary information.}
\label{tab:msvd_results}

\end{table*}

\begin{table*}[ht]
\centering
\resizebox{0.6\linewidth}{!}{%
\begin{tabular}{llcccccc}

&\multicolumn{1}{l}{Method}
&\multicolumn{1}{c}{Year}
&\multicolumn{1}{c}{Auxiliary Info$^{\dagger}$}
&\multicolumn{1}{c}{R@1$\uparrow$}
&\multicolumn{1}{c}{R@5$\uparrow$}
&\multicolumn{1}{c}{R@10$\uparrow$}
&\multicolumn{1}{c}{MR$\downarrow$}\\\hline

\multirow{5}{*}{\centering Zero-Shot (zs)}
&InternVideo2~\cite{DBLP:conf/eccv/WangLLYHCPZWSJLXZHQWW24}&2024&GMC&63.2&85.6&92.5&-\\\cline{2-8}
&GRAM~\cite{DBLP:conf/iclr/CicchettiGSC25}&2024&MM&59.0&-&91.1&-\\
&LanguageBind~\cite{DBLP:conf/iclr/ZhuLNYCWPJZLZ0024}&2024&MM&41.0&68.4&80.0&-\\\cline{2-8}

&IVR-QA~\cite{DBLP:conf/iccv/LiangA23}&2023&Q$\&$A&44.8&70.4&79.9&2\\
&MERLIN~\cite{DBLP:conf/emnlp/HanPLLK24}&2024&Q$\&$A&68.4&91.9&96.6&-\\\hline\hline

\multirow{20}{*}{\centering Fine-Tuning (ft)}
&HSE$^{*}$~\cite{DBLP:conf/eccv/ZhangHS18}&2018&VTC& 44.4 & 76.7 & 97.1 & 2 \\
&COOT~\cite{DBLP:conf/nips/GingZPB20}&2020&VTC&60.8&86.6&98.6&1\\\cline{2-8}

&CE~\cite{DBLP:conf/bmvc/LiuANZ19}&2019&VE&18.2&47.7&91.4&6\\
&MMT$^{*}$~\cite{DBLP:conf/eccv/Gabeur0AS20}&2020&VE&28.7&61.4&94.5&3\\
&T2VLAD~\cite{DBLP:conf/cvpr/WangZ021}&2021&VE&23.7&55.5&93.5&4\\
&DMMC$^{*}$~\cite{DBLP:conf/ijcai/WangZCCZPGWS21}&2021&VE&30.2&63.5&95.3&3\\
&CMGSD~\cite{DBLP:conf/sigir/HeWFJLZT21}&2021&VE& 24.2 & 56.3 & 94.0& 4\\ 
&HCQ~\cite{DBLP:conf/www/WangCLZLXX22}&2022&VE&22.2& 53.7 & 91.2 & 5\\\cline{2-8}

&HiT$^{*}$~\cite{DBLP:conf/iccv/Liu0QCDW21}&2021&Audio&29.6&60.7&95.6&3\\
&MMC$^{*}$~\cite{DBLP:conf/wacv/GabeurN0AS22}&2022&Audio& 29.0& 61.7& 94.6& 4\\
&EclipSE~\cite{DBLP:conf/eccv/LinLBB22}&2022&Audio& 42.3 & 73.2 & 83.8 & -\\\cline{2-8}

&Support-Set$^{*}$~\cite{DBLP:conf/iclr/Patrick0AMHHV21}&2021&TC& 29.2&61.6&84.7&3\\\cline{2-8}

&TeachText~\cite{DBLP:conf/iccv/CroitoruBLJZAL21}&2021&TE& 23.5& 57.2& 96.1& 4\\\cline{2-8}

&TACo$^{*}$~\cite{DBLP:conf/iccv/YangBG21}&2021&POS& 30.4& 61.2& 93.4& 3\\\cline{2-8}

&VAST$^{*}$~\cite{DBLP:conf/nips/ChenLWZSZL23}&2023&GMC&70.5&90.9&95.5&-\\\cline{2-8}

&GRAM$^{*}$~\cite{DBLP:conf/iclr/CicchettiGSC25}&2024&MM&69.9&-&96.1&-\\\cline{2-8}

&CLIP-ViP$^{*}$~\cite{DBLP:conf/iclr/XueS0FSLL23}&2023&GVC& 51.1&78.4&88.3&1\\
&M-RAAP$^{*}$~\cite{DBLP:conf/sigir/DongFZYYG24}&2024&GVC& 41.2&70.2&82.4&-\\
&T2VIndexer$^{*}$~\cite{DBLP:conf/mm/Li0GL0024}&2024&GVC& 54.9&82.5&90.0&-\\\cline{2-8}

&SHE-Net~\cite{DBLP:journals/corr/abs-2404-14066}&2024&HD&43.9& 75.3& 86.1& 2\\\hline\hline

\end{tabular}}
\caption{Paragraph-to-Video results on ActivityNet Captions. $^{*}$: Pre-training. $^{\dagger}$: see Tabs~\ref{tab:explanation_auxiliary_video} -~\ref{tab:explanation_auxiliary_text} for the type of auxiliary information.}
\label{tab:activity_results}

\end{table*}

\begin{table*}[ht]
\centering
\resizebox{0.6\linewidth}{!}{%
\begin{tabular}{llcccccc}

&\multicolumn{1}{l}{Method}
&\multicolumn{1}{c}{Year}
&\multicolumn{1}{c}{Auxiliary Info$^{\dagger}$}
&\multicolumn{1}{c}{R@1$\uparrow$}
&\multicolumn{1}{c}{R@5$\uparrow$}
&\multicolumn{1}{c}{R@10$\uparrow$}
&\multicolumn{1}{c}{MR$\downarrow$}\\\hline

\multirow{9}{*}{\centering Zero-Shot (zs)}
&OA-Trans~\cite{DBLP:conf/cvpr/WangGCY0SQS22}&2022&ROI&23.5&50.4&59.8&6\\\cline{2-8}
&ALPRO~\cite{DBLP:conf/cvpr/Li0LNH22}&2022&VT& 23.8&47.3&57.9&6\\\cline{2-8}
&BridgeFormer~\cite{DBLP:conf/cvpr/GeGLLSQL22}&2022&POS&25.6&50.6&61.1&5\\\cline{2-8}
&IVR-QA~\cite{DBLP:conf/iccv/LiangA23}&2023&Q$\&$A&61.6&86.9&91.5&1\\\cline{2-8}
&In-Style~\cite{DBLP:conf/iccv/ShvetsovaKSK23}&2023&GVC&29.4&59.2&68.6&3\\\cline{2-8}

&VAST~\cite{DBLP:conf/nips/ChenLWZSZL23}&2023&GMC&55.5&74.3&79.6&-\\
&InternVideo2~\cite{DBLP:conf/eccv/WangLLYHCPZWSJLXZHQWW24}&2024&GMC&57.9&80.0&84.6&-\\\cline{2-8}

&GRAM~\cite{DBLP:conf/iclr/CicchettiGSC25}&2024&MM&54.2&-&79.3&-\\
&LanguageBind~\cite{DBLP:conf/iclr/ZhuLNYCWPJZLZ0024}&2024&MM&39.9&66.1&74.6&-\\\hline\hline

\multirow{19}{*}{\centering Fine-Tuning (ft)}
&HSE$^{*}$~\cite{DBLP:conf/eccv/ZhangHS18}&2018&VTC& 29.7&60.3&92.4&3\\\cline{2-8}

&CE~\cite{DBLP:conf/bmvc/LiuANZ19}&2019&VE& 16.1&41.1&82.7&8\\\cline{2-8}

&TeachText~\cite{DBLP:conf/iccv/CroitoruBLJZAL21}&2021&TE&21.6&48.6&62.9&6\\\cline{2-8}

&OA-Trans$^{*}$~\cite{DBLP:conf/cvpr/WangGCY0SQS22}&2022&ROI&34.8&64.4&75.1&3\\
&MSIA~\cite{chen2024multilevel}&2024&ROI&43.6&70.2&79.6&2\\\cline{2-8}

&ALPRO$^{*}$~\cite{DBLP:conf/cvpr/Li0LNH22}&2022&VT& 35.9&67.5&78.8&3\\
&HiSE~\cite{DBLP:conf/mm/WangXHLJHD22}&2022&VT& 44.1&69.9&80.3&2\\
&TABLE$^{*}$~\cite{DBLP:conf/aaai/ChenWLQMS23}&2023&VT& 47.9&74.0&82.1&2\\\cline{2-8}

&EclipSE~\cite{DBLP:conf/eccv/LinLBB22}&2022&Audio&44.2&-&-&-\\\cline{2-8}

&BridgeFormer$^{*}$~\cite{DBLP:conf/cvpr/GeGLLSQL22}&2022&POS&37.0&62.2&73.9&3\\
&WAVER~\cite{DBLP:conf/icassp/LeK0L24}&2024&POS&49.2&77.2&85.6&2\\\cline{2-8}

&VAST$^{*}$~\cite{DBLP:conf/nips/ChenLWZSZL23}&2023&GMC&72.0&89.0&91.4&-\\\cline{2-8}

&GRAM$^{*}$~\cite{DBLP:conf/iclr/CicchettiGSC25}&2024&MM&67.3&-&90.1&-\\\cline{2-8}

&CLIP-ViP$^{*}$~\cite{DBLP:conf/iclr/XueS0FSLL23}&2023&GVC&48.6&77.1&84.4&2\\
&Cap4Video~\cite{DBLP:conf/cvpr/WuLFWO23}&2023&GVC&52.0&79.4&87.5&1\\
&M-RAAP$^{*}$~\cite{DBLP:conf/sigir/DongFZYYG24}&2024&GVC&43.6&68.8&77.3&-\\
&T2VIndexer$^{*}$~\cite{DBLP:conf/mm/Li0GL0024}&2024&GVC&51.9&79.2&87.1&-\\\cline{2-8}

&MuMUR~\cite{DBLP:journals/ir/MadasuASRTBL23}&2023&L&44.4&74.3&83.1&2\\\cline{2-8}

&SHE-Net~\cite{DBLP:journals/corr/abs-2404-14066}&2024&HD&45.6&75.6&83.6&2\\\hline\hline

\end{tabular}}
\caption{Paragraph-to-Video results on DiDeMo. $^{*}$: Pre-training. $^{\dagger}$: see Tabs~\ref{tab:explanation_auxiliary_video} -~\ref{tab:explanation_auxiliary_text} for the type of auxiliary information.}
\label{tab:didemo_results}

\end{table*}

\begin{table*}[ht]
\centering
\resizebox{0.6\linewidth}{!}{%
\begin{tabular}{llcccccc}

&\multicolumn{1}{l}{Method}
&\multicolumn{1}{c}{Year}
&\multicolumn{1}{c}{Auxiliary Info$^{\dagger}$}
&\multicolumn{1}{c}{R@1$\uparrow$}
&\multicolumn{1}{c}{R@5$\uparrow$}
&\multicolumn{1}{c}{R@10$\uparrow$}
&\multicolumn{1}{c}{MR$\downarrow$}\\\hline

\multirow{6}{*}{\centering Zero-Shot (zs)}
&AVLnet~\cite{DBLP:conf/interspeech/RouditchenkoBHC21}&2021&Audio&1.4&5.9&9.4&273\\\cline{2-8}
&BridgeFormer~\cite{DBLP:conf/cvpr/GeGLLSQL22}&2022&POS&12.2&25.9&32.2&40\\\cline{2-8}
&IVR-QA~\cite{DBLP:conf/iccv/LiangA23}&2023&Q$\&$A&55.7&78.0&85.9&1\\\cline{2-8}

&InternVideo2~\cite{DBLP:conf/eccv/WangLLYHCPZWSJLXZHQWW24}&2024&GMC&33.8&55.9&62.2&-\\\cline{2-8}

&In-Style~\cite{DBLP:conf/iccv/ShvetsovaKSK23}&2023&GVC&16.4&30.1&38.7&28\\\cline{2-8}
&HowToCaption~\cite{DBLP:conf/eccv/ShvetsovaKHRSK24}&2023&GTC&17.3&31.7&38.6&29\\\hline\hline

\multirow{27}{*}{\centering Fine-Tuning (ft)}
&CT-SAN~\cite{DBLP:conf/cvpr/YuKCK17}&2017&POS&5.1&16.3&25.2&46\\
&CAMoE~\cite{DBLP:journals/corr/abs-2109-04290}&2021&POS&22.5&42.6&50.9&-\\
&BridgeFormer$^{*}$~\cite{DBLP:conf/cvpr/GeGLLSQL22}&2022&POS&17.9&35.4&44.5&15\\\cline{2-8}

&MoE~\cite{DBLP:journals/corr/abs-1804-02516}&2018&VE&10.1&25.6&34.6&27\\
&CE~\cite{DBLP:conf/bmvc/LiuANZ19}&2019&VE&11.2&26.9&34.8&25\\
&MMT$^{*}$~\cite{DBLP:conf/eccv/Gabeur0AS20}&2020&VE&12.9&29.9&40.1&19\\
&T2VLAD~\cite{DBLP:conf/cvpr/WangZ021}&2021&VE&14.3&32.4&42.2&16\\
&CMGSD~\cite{DBLP:conf/sigir/HeWFJLZT21}&2021&VE&14.0&31.1&41.0&18\\
&MDMMT~\cite{DBLP:conf/cvpr/DzabraevKKP21}&2021&VE& 18.8&38.5&47.9&12\\
&DMMC$^{*}$~\cite{DBLP:conf/ijcai/WangZCCZPGWS21}&2021&VE&15.8&34.1&43.6&14\\
&HCQ~\cite{DBLP:conf/www/WangCLZLXX22}&2022&VE&14.5&33.6&43.1&18\\
&MDMMT-2$^{*}$~\cite{DBLP:journals/corr/abs-2203-07086}&2022&VE&26.9&46.7&55.9&7\\\cline{2-8}

&JSFusion~\cite{DBLP:conf/eccv/YuKK18}&2018&Audio&9.1&21.2&34.1&36\\
&AVLnet$^{*}$~\cite{DBLP:conf/interspeech/RouditchenkoBHC21}&2021&Audio&11.4&26.0&34.6&30\\
&HiT~\cite{DBLP:conf/iccv/Liu0QCDW21}&2021 &Audio&14.0&31.2&41.6&18\\
&TEFAL~\cite{DBLP:conf/iccv/IbrahimiSWGSO23}&2023&Audio&26.8&46.1&56.5&7\\\cline{2-8}

&TCE~\cite{DBLP:conf/sigir/YangD0W0C20}&2020&HD&10.6&25.8&35.1&29\\\cline{2-8}

&TeachText~\cite{DBLP:conf/iccv/CroitoruBLJZAL21}&2021&TE&17.2&36.5&46.3&14\\\cline{2-8}

&OA-Trans$^{*}$~\cite{DBLP:conf/cvpr/WangGCY0SQS22}&2022&ROI&18.2&34.3&43.7&18\\\cline{2-8}

&STG~\cite{DBLP:journals/tmm/SongCWJ22}&2022&STD&10.3&23.1&31.7&35\\
&MGM~\cite{DBLP:conf/mm/SongCJ23}&2023&STD&24.0&44.0&54.4&8\\\cline{2-8}

&TABLE$^{*}$~\cite{DBLP:conf/aaai/ChenWLQMS23}&2023&VT&24.3&44.9&53.7&8\\\cline{2-8}

&CLIP-ViP$^{*}$~\cite{DBLP:conf/iclr/XueS0FSLL23}&2023&GVC&25.6&45.3&54.4&8\\
&M-RAAP$^{*}$~\cite{DBLP:conf/sigir/DongFZYYG24}&2024&GVC&24.4&44.3&52.0&-\\\cline{2-8}

&MSIA~\cite{chen2024multilevel}&2024&ROI&19.7&38.1&47.5&12\\\cline{2-8}

&MT-AGCN~\cite{DBLP:journals/mms/LvSN24}&2024&HD&25.9&45.1&54.3&8\\\cline{2-8}

&GQE~\cite{bai2025bridging}&2024&GTC&28.3&49.2&59.1&6\\\hline\hline
\end{tabular}}
\caption{Sentence-to-Clip results on LSMDC. $^{*}$: Pre-training. $^{\dagger}$: see Tabs~\ref{tab:explanation_auxiliary_video} -~\ref{tab:explanation_auxiliary_text} for the type of auxiliary information.}
\label{tab:lsmdc_results}

\end{table*}

\begin{table*}[ht]
\centering
\begin{minipage}[t]{0.49\textwidth}
\centering
\resizebox{\linewidth}{!}{%
\begin{tabular}{llcccccc}

&\multicolumn{1}{l}{Method}
&\multicolumn{1}{c}{Year}
&\multicolumn{1}{c}{Auxiliary Info$^{\dagger}$}
&\multicolumn{1}{c}{R@1$\uparrow$}
&\multicolumn{1}{c}{R@5$\uparrow$}
&\multicolumn{1}{c}{R@10$\uparrow$}
&\multicolumn{1}{c}{MR$\downarrow$}\\\hline

\multirow{9}{*}{\centering Zero-Shot (zs)}
&ActBERT~\cite{DBLP:conf/cvpr/ZhuY20a}&2020&ROI&9.6&26.7&38.0&19\\\cline{2-8}
&MIL-NCE~\cite{DBLP:conf/cvpr/MiechASLSZ20}&2020&TD&15.1&38.0&51.2&10\\\cline{2-8}

&MCN~\cite{DBLP:conf/iccv/ChenRDK0BPKFHGP21}&2021&Audio&18.1&35.5&45.2&-\\
&AVLnet~\cite{DBLP:conf/interspeech/RouditchenkoBHC21}&2021&Audio&19.9&36.1&44.3&16\\
&VATT~\cite{DBLP:conf/nips/AkbariYQCCCG21}&2021&Audio& -&-&45.5&13\\
&EAO~\cite{DBLP:conf/cvpr/ShvetsovaCR0KFH22}&2022&Audio& 24.6& 48.3&60.4&6\\\cline{2-8}

&TACo~\cite{DBLP:conf/iccv/YangBG21}&2021&POS&19.9&43.2&55.7&8\\\cline{2-8}

&In-Style~\cite{DBLP:conf/iccv/ShvetsovaKSK23}&2023&GVC&6.8&16.7&24.5&63\\\cline{2-8}

&HowToCaption~\cite{DBLP:conf/eccv/ShvetsovaKHRSK24}&2023&GTC&13.4&33.1&44.1&15\\\hline\hline

\multirow{10}{*}{\centering Fine-Tuning (ft)}
&MDMMT-2$^{*}$~\cite{DBLP:journals/corr/abs-2203-07086}&2022&VE&32.0&64.0&74.8&3\\\cline{2-8}

&COOT~\cite{DBLP:conf/nips/GingZPB20}&2020&VTC& 16.7 & 40.2 & 52.3 & 9\\
&ConTra~\cite{DBLP:conf/accv/FragomeniWD22}&2022&VTC& 16.7& 42.1& 55.2& 8\\\cline{2-8}

&AVLnet$^{*}$~\cite{DBLP:conf/interspeech/RouditchenkoBHC21}&2021&Audio&30.2&55.5&66.5&4\\
&MMC$^{*}$~\cite{DBLP:conf/wacv/GabeurN0AS22}&2022&Audio& 23.2&48.0&58.6&6\\
&EAO$^{*}$~\cite{DBLP:conf/cvpr/ShvetsovaCR0KFH22}&2022&Audio&32.1& 59.1&70.9&3\\\cline{2-8}

&FCA-Net~\cite{DBLP:conf/mm/HanCXZZC21}&2021&POS&12.2&31.5&42.6&14\\
&TACo$^{*}$~\cite{DBLP:conf/iccv/YangBG21}&2021&POS&29.6&59.7&72.7&4\\\cline{2-8}

&VAST$^{*}$~\cite{DBLP:conf/nips/ChenLWZSZL23}&2023&GMC&50.4&74.3&80.8&-\\\cline{2-8}

&C2KD~\cite{DBLP:conf/icassp/RouditchenkoCSTFKKHKG23}&2023&L&15.5&-&-&-\\\hline\hline
\end{tabular}}
\caption{Sentence-to-Clip results on YouCook2. $^{*}$: Pre-training. $^{\dagger}$: see Tabs~\ref{tab:explanation_auxiliary_video} -~\ref{tab:explanation_auxiliary_text} for the type of auxiliary information.}
\label{tab:yc2_results}

\end{minipage}
\hfill
\begin{minipage}[t]{0.49\textwidth}
\centering
\resizebox{\linewidth}{!}{%
\begin{tabular}{llcccccc}

&\multicolumn{1}{l}{Method}
&\multicolumn{1}{c}{Year}
&\multicolumn{1}{c}{Auxiliary Info$^{\dagger}$}
&\multicolumn{1}{c}{R@1$\uparrow$}
&\multicolumn{1}{c}{R@5$\uparrow$}
&\multicolumn{1}{c}{R@10$\uparrow$}
&\multicolumn{1}{c}{MR$\downarrow$}\\\hline

\multirow{3}{*}{\centering Zero-Shot (zs)}
&InternVideo2~\cite{DBLP:conf/eccv/WangLLYHCPZWSJLXZHQWW24}&2024&GMC&71.5&94.0&97.1&-\\\cline{2-8}

&GRAM~\cite{DBLP:conf/iclr/CicchettiGSC25}&2024&MM&83.5&-&98.8&-\\\cline{2-8}

&VIIT~\cite{DBLP:conf/cvpr/00060ZWCMSA0GKY24}&2024&GVC&65.6&91.7&95.8&-\\\hline\hline

\multirow{17}{*}{\centering Fine-Tuning (ft)}
&LAFF$^{*}$~\cite{DBLP:conf/eccv/HuCWZDL22}&2022&VE&59.1&91.7&96.3&-\\\cline{2-8}

&HGR~\cite{DBLP:conf/cvpr/ChenZJW20}&2020&HD&35.1&73.5&83.5&2\\
&HANeT~\cite{DBLP:conf/mm/WuHTLL21}&2021&HD&36.4&74.1&84.1&2\\\cline{2-8}

&Support-Set$^{*}$~\cite{DBLP:conf/iclr/Patrick0AMHHV21}&2021&TC&45.9&82.4&90.4&1\\
&LINAS~\cite{DBLP:conf/eccv/FangWZHH22}&2022&TC&-&74.4&-&2\\\cline{2-8}

&TeachText~\cite{DBLP:conf/iccv/CroitoruBLJZAL21}&2021&TE&53.2&87.4&93.3&1\\\cline{2-8}

&MMP$^{*}$~\cite{DBLP:conf/naacl/HuangPHNMH21}&2021&L&44.3&80.7&88.9&-\\\cline{2-8}

&FCA-Net~\cite{DBLP:conf/mm/HanCXZZC21}&2021&POS&41.3&82.1&91.2&1\\
&Dual Encoding~\cite{DBLP:journals/pami/DongLXYYWW22}&2022&POS&36.8&73.6&83.7&-\\
&WAVER~\cite{DBLP:conf/icassp/LeK0L24}&2024&POS&66.5&93.3&97.0&1\\\cline{2-8}

&CLIP-bnl~\cite{DBLP:conf/mm/WangCH022}&2022&POS-C&57.6&88.3&94.0&-\\\cline{2-8}

&TEFAL~\cite{DBLP:conf/iccv/IbrahimiSWGSO23}&2023&Audio&61.0&90.4&95.3&1\\\cline{2-8}

&Cap4Video~\cite{DBLP:conf/cvpr/WuLFWO23}&2023&GVC&66.6&93.1&97.0&1\\
&M-RAAP$^{*}$~\cite{DBLP:conf/sigir/DongFZYYG24}&2024&GVC&58.0&86.3&92.6&-\\\cline{2-8}

&VAST$^{*}$~\cite{DBLP:conf/nips/ChenLWZSZL23}&2023&GMC&83.0&98.2&99.2&-\\\cline{2-8}

&GRAM$^{*}$~\cite{DBLP:conf/iclr/CicchettiGSC25}&2024&MM&87.7&-&100.0&-\\\cline{2-8}

&GQE~\cite{bai2025bridging}&2024&GTC&63.3&92.4&96.8&1\\\hline\hline

\end{tabular}}
\caption{Sentence-to-Clip results on VATEX. $^{*}$: Pre-training. $^{\dagger}$: see Tabs~\ref{tab:explanation_auxiliary_video} -~\ref{tab:explanation_auxiliary_text} for the type of auxiliary information.}
\label{tab:vatex_results}
\end{minipage}
\end{table*}

\begin{table*}[ht]
\centering
\resizebox{0.6\linewidth}{!}{%
\begin{tabular}{llcccc}

&\multicolumn{1}{l}{Method}
&\multicolumn{1}{c}{Year}
&\multicolumn{1}{c}{Auxiliary Info$^{\dagger}$}
&\multicolumn{1}{c}{mAP$\uparrow$}
&\multicolumn{1}{c}{nDCG$\uparrow$}\\\hline

\multirow{3}{*}{\centering Zero-Shot (zs)}
&HH~\cite{DBLP:conf/iccv/Zhang0Z23a}&2023&STD&32.7&36.2\\\cline{2-6}
&EgoNCE++~\cite{DBLP:conf/iclr/XuWDSZJ25}&2024&GTC&32.8&33.6\\\cline{2-6}
&LaViLa~\cite{DBLP:conf/cvpr/0006MKG23}&2023&GVC&32.2&33.2\\\hline\hline


\multirow{2}{*}{\centering Fine-Tuning (ft)}
&JPoSE~\cite{DBLP:conf/iccv/WrayCLD19}&2019&POS&38.1&51.6\\\cline{2-6}
&LaViLa$^{*}$~\cite{DBLP:conf/cvpr/0006MKG23}&2023&GVC&47.1&64.9\\\hline\hline

\end{tabular}}
\caption{Sentence-to-Clip results on EPIC-KITCHENS-100. $^{*}$: Pre-training. $^{\dagger}$: see Tabs~\ref{tab:explanation_auxiliary_video} -~\ref{tab:explanation_auxiliary_text} for the type of auxiliary information.}
\label{tab:epic_results}

\end{table*}

\end{document}